\documentclass{article}

\PassOptionsToPackage{numbers, compress}{natbib}
 \usepackage[preprint]{neurips_2026}

\usepackage[utf8]{inputenc} 
\usepackage[T1]{fontenc}    
\usepackage{hyperref}       
\usepackage{url}            
\usepackage{booktabs}       
\usepackage{nicefrac}       
\usepackage{amsmath,amsfonts,bm}
\usepackage{fontawesome5}
\usepackage{twemojis}
\usepackage{comment}
\usepackage{xspace}
\usepackage{graphicx}
\usepackage{mwe}
\usepackage{subcaption}
\usepackage{pdflscape}
\usepackage{adjustbox} 
\usepackage{colortbl}
\usepackage[dvipsnames]{xcolor}
\usepackage{pifont}
\usepackage{wrapfig}
\usepackage{enumitem}
\usepackage{adjustbox}
\usepackage{longtable}
\usepackage{svg}
\usepackage{makecell}
\usepackage{algorithm}
\usepackage{colortbl}
\usepackage{amssymb}
\usepackage{algpseudocode}
\usepackage{xstring}
\usepackage{tabularx}
\usepackage{siunitx}
\usepackage[super]{nth}
\usepackage[noabbrev, capitalise]{cleveref}
\usepackage[framemethod=tikz]{mdframed}

\newcommand{\ours}{3D-SC\xspace}
\newcommand{\oursLong}{Geometry Matters: 3D Foundation Priors for Learning Semantic Correspondence}

\usepackage{microtype}
\renewcommand{\paragraph}[1]{\noindent\textbf{#1\IfEndWith{#1}{?}{}{.}}}
\newcommand{\paragraphit}[1]{\noindent\textit{#1\IfEndWith{#1}{?}{}{.}}}
\makeatletter
\DeclareRobustCommand\onedot{\futurelet\@let@token\@onedot}
\def\@onedot{\ifx\@let@token.\else.\null\fi\xspace}

\def\eg{\emph{e.g}\onedot} 

\def\ie{\emph{i.e}\onedot}

\makeatother

\newcommand\pckTen{$\text{PCK@}0.1$}

\newcommand{\myrowcolour}{\rowcolor[gray]{0.925}}

\newcommand{\Shape}{S}
\newcommand{\PointPixel}{p}
\newcommand{\PointThreeD}{\mathbf{v}}
\newcommand{\Feature}{\mathcal{F}}
\newcommand{\Mesh}{\mathcal{M}}
\newcommand{\SetPixelPoint}{\mathcal{P}}

\newcommand{\Bench}[1][]{\includegraphics[height=10pt]{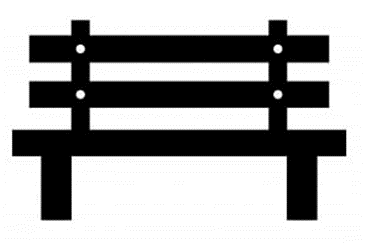}}
\newcommand{\Microwave}[1][]{\includegraphics[height=10pt]{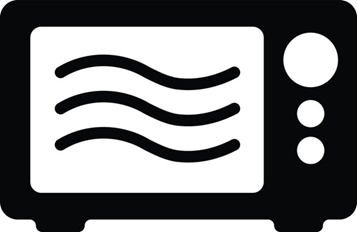}}
\newcommand{\Mouse}[1][]{\includegraphics[height=10pt]{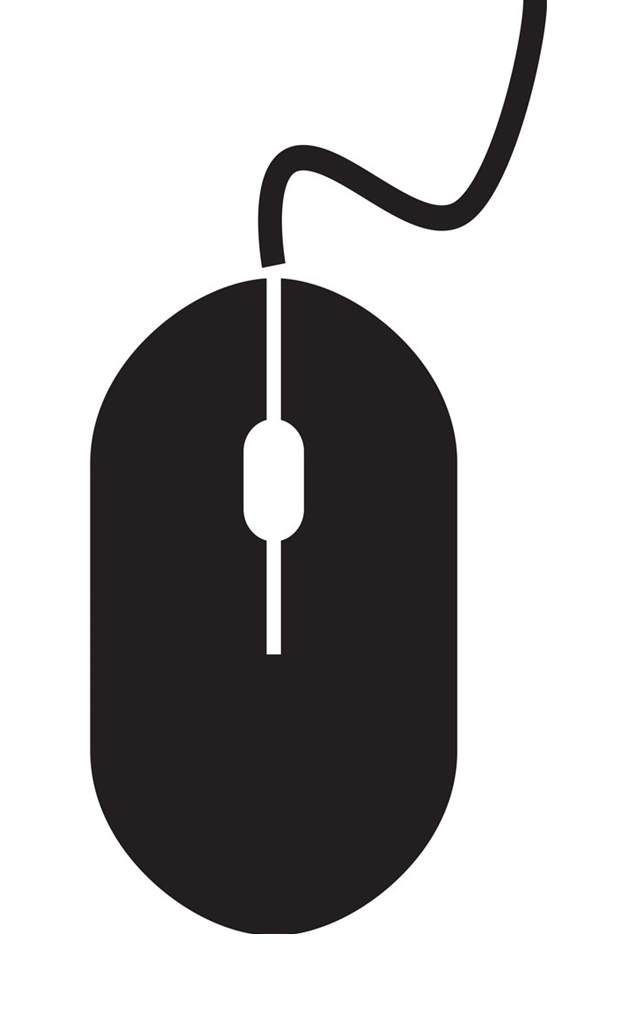}}
\newcommand{\Remote}[1][]{\includegraphics[height=10pt]{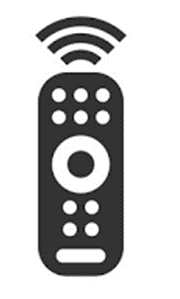}}
\newcommand{\Toaster}[1][]{\includegraphics[height=10pt]{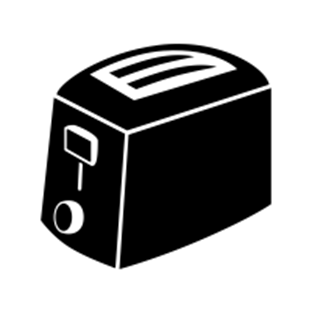}}
\newcommand{\Toilet}[1][]{\includegraphics[height=10pt]{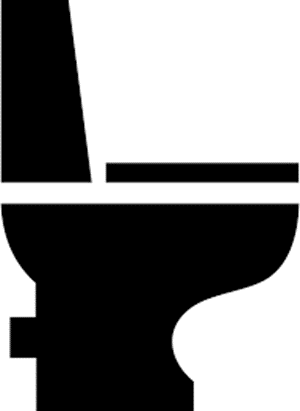}}
\newcommand{\Trident}[1][]{\includegraphics[height=12pt]{figures/icons/trident.png}}
\newcommand{\Cow}[1][]{\includegraphics[width=10pt,trim={6cm 9cm 5cm 6cm},clip]{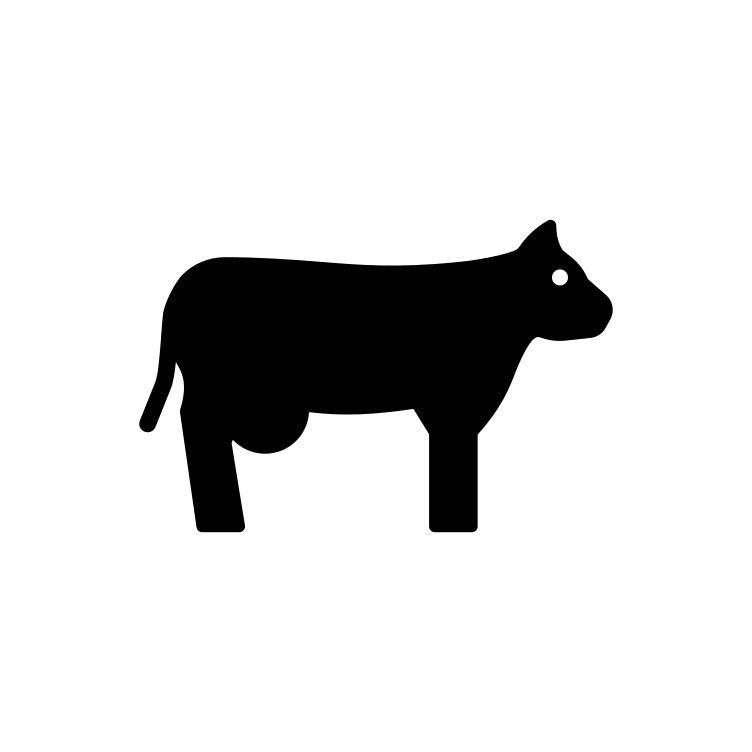}}
\newcommand{\Plant}[1][]{\includegraphics[width=10pt,trim={7cm 6cm 5cm 2cm},clip]{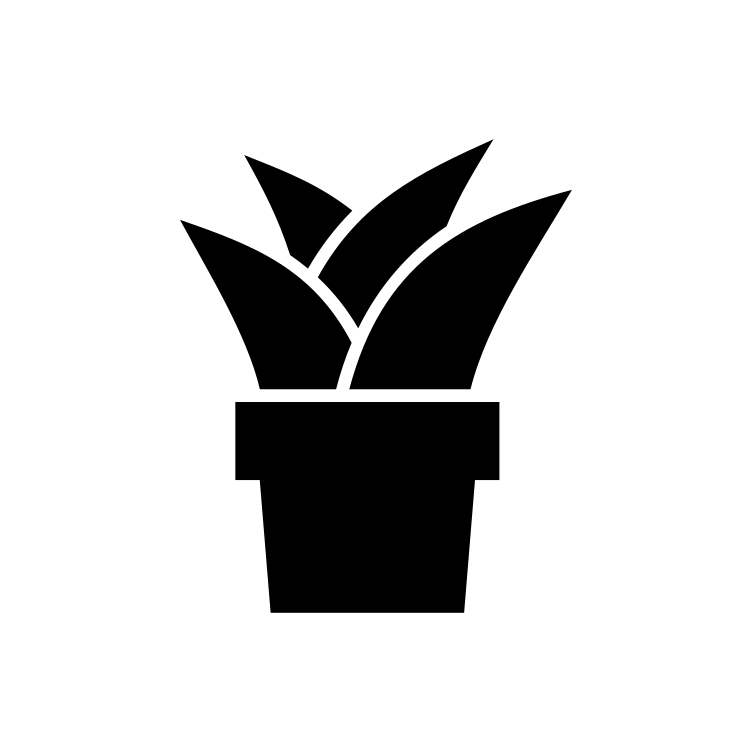}}
\newcommand{\Sheep}[1][]{\includegraphics[width=10pt,trim={6cm 7cm 5cm 6cm},clip]{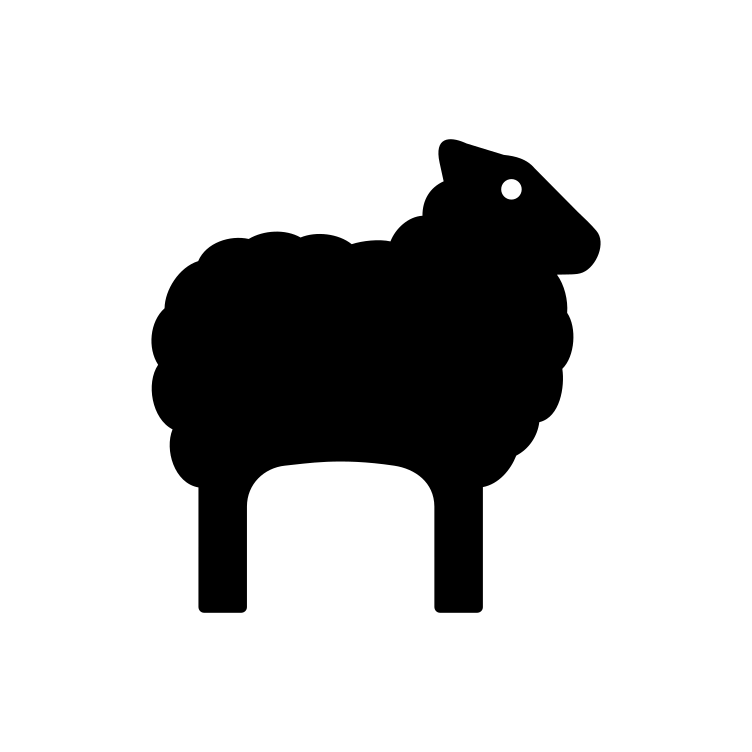}}

\definecolor{pipelinepurple}{RGB}{150,115,166}
\definecolor{pipelineorange}{RGB}{255, 128, 0}

\title{\oursLong}

\author{%
  Artur Jesslen$^{1}$ \quad
  Olaf D\"unkel$^{2}$ \quad
  Adam Kortylewski$^{3}$
  \\
  $^{1}$University of Freiburg, Germany \\
  $^{2}$Max Planck Institute for Informatics, Saarland Informatics Campus, Germany \\
  $^{3}$CISPA Helmholtz Center for Information Security, Germany
}

\begin{document}

\maketitle

\vspace{-0.6em}
\begin{abstract}
Foundation features from self-supervised vision models and text-to-image diffusion models have proven effective for semantic correspondence estimation. However, because these features are learned primarily from 2D image objectives, they lack explicit 3D awareness and often confuse symmetric object sides, repeated parts, and visually similar structures that are distinct in 3D. We introduce a 3D-aware post-training framework that goes beyond available 2D foundation features by incorporating priors from 3D foundation models. Given an image, our method uses SAM3D to estimate object geometry and pose, and refines the pose through render-and-compare optimization. Subsequently, we render PartField descriptors from the reconstructed geometry into the image plane based on the estimated object pose. The resulting geometry-aware feature maps complement DINO and Stable Diffusion features, while geodesic distances on the reconstructed shapes enable reliable filtering of candidate correspondences. We use the filtered matches as supervision to train a lightweight adapter on top of DINO and Stable Diffusion for semantic correspondence. In contrast to prior post-training approaches that require pose annotations and rely on coarse spherical geometry, our method automatically obtains instance-specific 3D structure and uses it to guide correspondence learning. Experiments show that our approach improves semantic correspondence over the prior methods while reducing manual geometric supervision. Code and model can be found at 
\href{https://github.com/GenIntel/3D-SC}{\faGithub\texttt{/GenIntel/3D-SC}}.
\end{abstract}

\begin{figure}[h]
    \vspace{-1em}
    \centering

    \begin{minipage}[t]{0.32\linewidth}
        \centering
        \includegraphics[width=\linewidth]{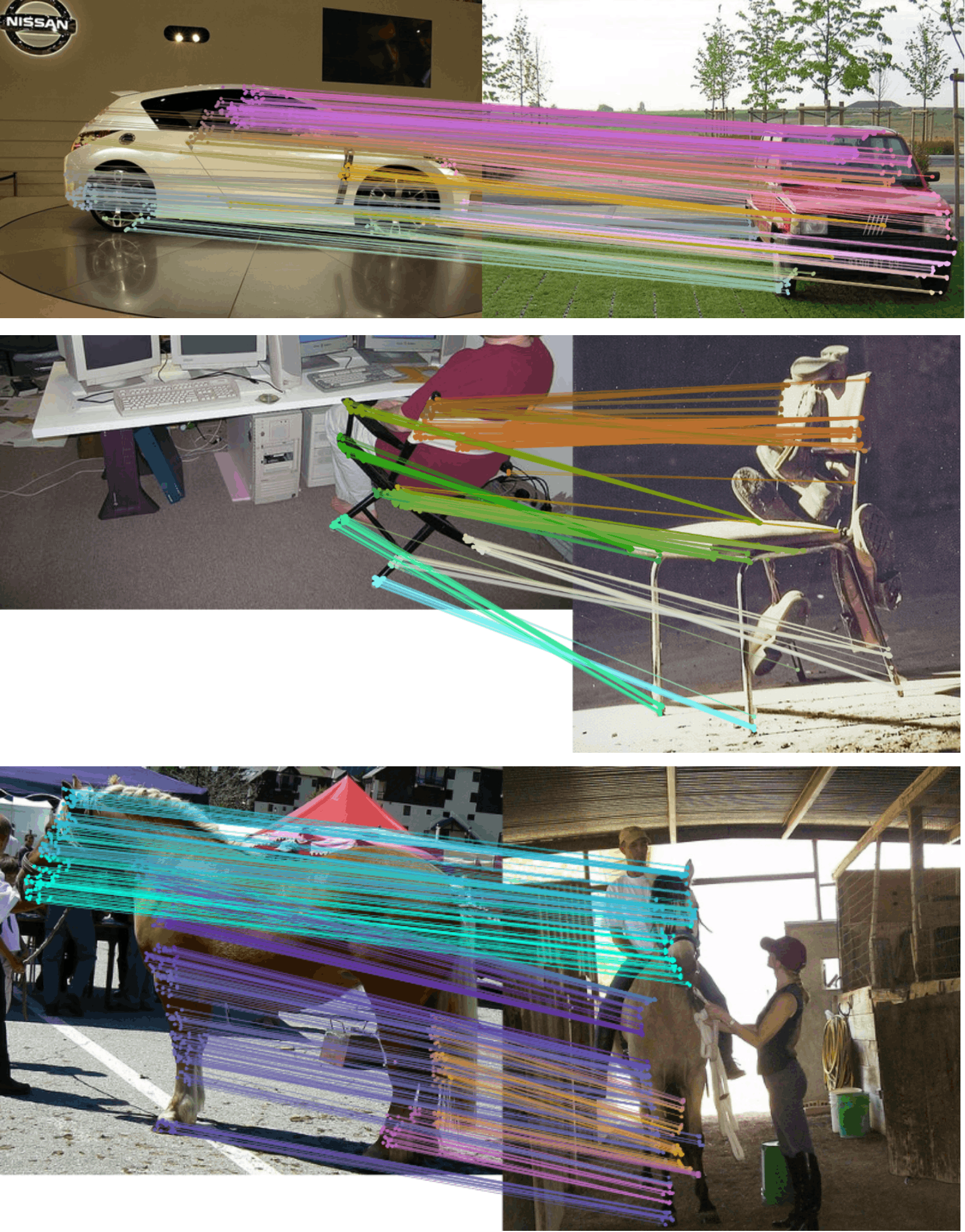}
        \subcaption{\textbf{SD+DINO.}}
    \end{minipage}
    \hfill
    \begin{minipage}[t]{0.32\linewidth}
        \centering
        \includegraphics[width=\linewidth]{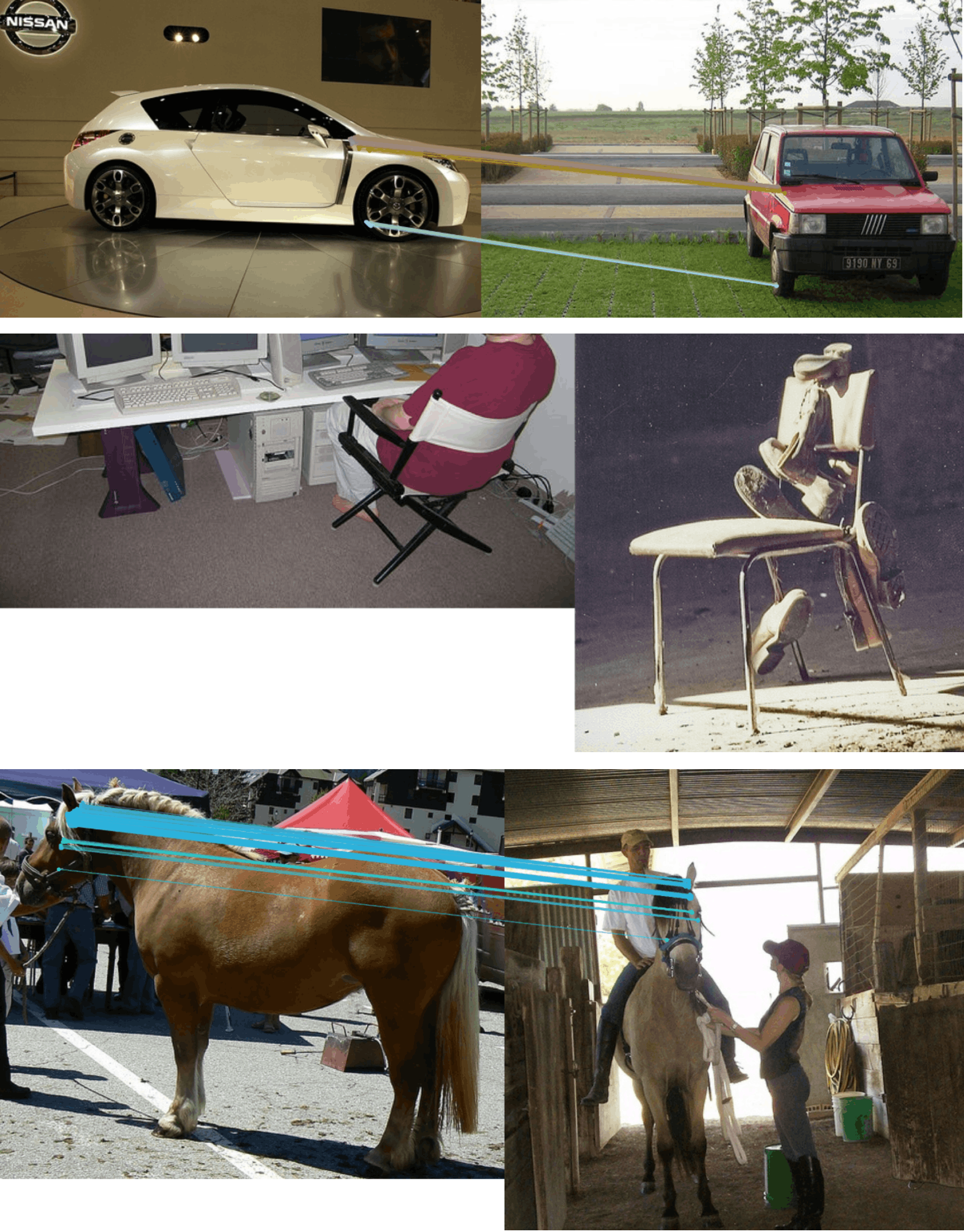}
        \subcaption{\centering\textbf{SD+DINO\\+ Geodesic Filtering.}}
    \end{minipage}
    \hfill
    \begin{minipage}[t]{0.32\linewidth}
        \centering
        \includegraphics[width=\linewidth]{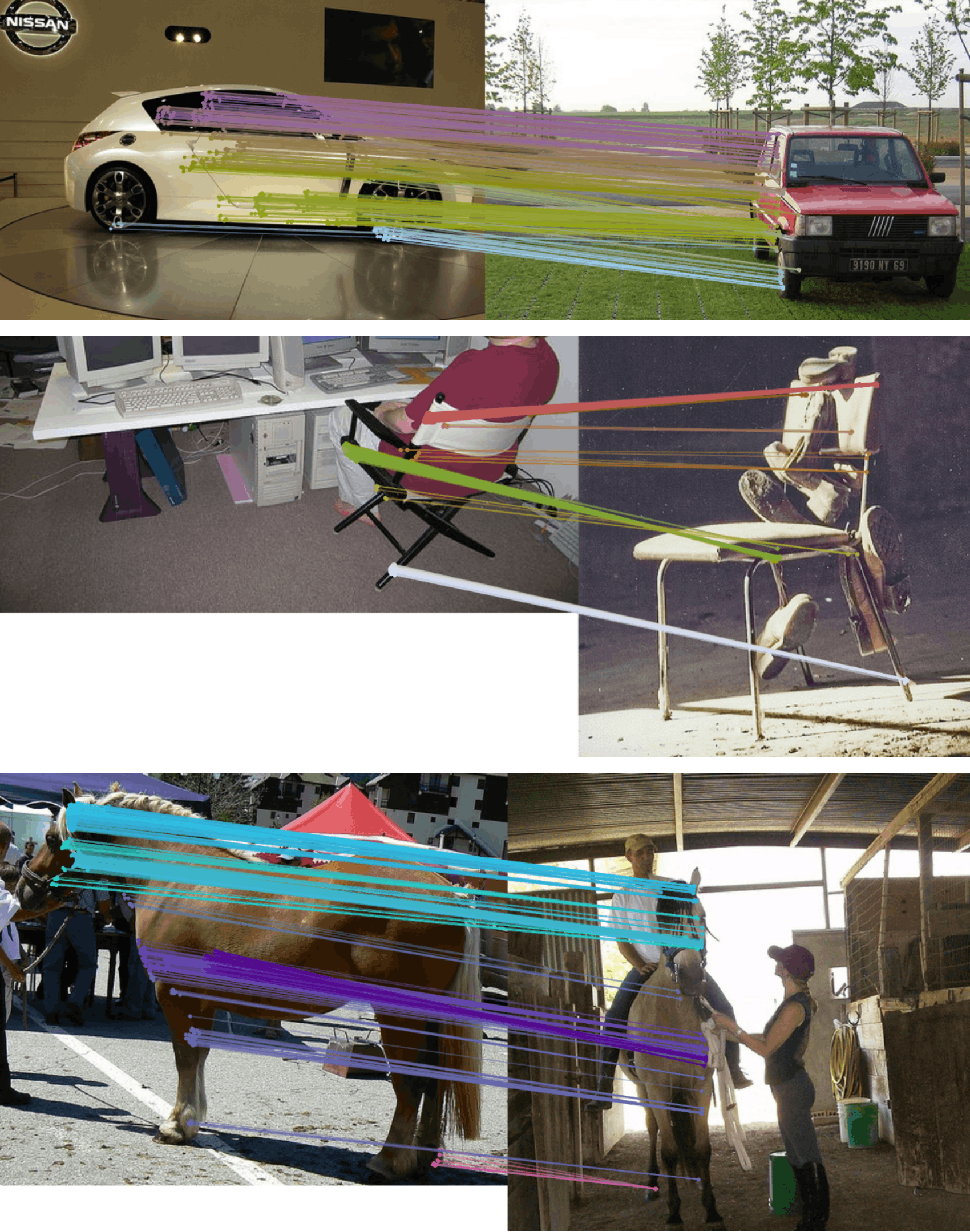}
        \subcaption{\centering\textbf{SD+DINO+Partfield\\+ Geodesic Filtering.}}
    \end{minipage}

    \caption{\textbf{3D foundation priors improve both candidate generation and filtering of semantic correspondences.} Existing zero-shot pipelines based on SD+DINO (a) suffer from left--right and repeated-part confusion, producing many incorrect matches. Adding our geodesic filter (b) removes wrong matches but is bottlenecked by feature quality, often leaving few surviving correspondences. Adding PartField features (c) yields dense and accurate correspondences even with large pose changes.}
    \label{fig:teaser}
    \vspace{-1em}
\end{figure}

\section{Introduction}
\label{sec:intro}

Semantic correspondence aims to establish matches between semantically equivalent object parts across different images and is a fundamental problem in visual recognition, with many applications like in vision~\citep{wang2024gs} or robotics~\citep{zhu2024densematcher}.
Unlike low-level image matching, semantic correspondence requires robustness to changes in appearance, viewpoint, articulation, intra-class shape variation, and background clutter.
As a result, it remains challenging to match object parts that are visually different but semantically equivalent, or visually similar but semantically distinct.

Recent progress has been driven by foundation features, with self-supervised vision transformers (DINOv2) and text-to-image diffusion models (Stable Diffusion) producing representations that transfer surprisingly well to dense semantic matching~\citep{caron2021emerging,amir2022deep,oquab2023dinov2,rombach2022high,tumanyan2023plug}.
Their fusion has become a strong zero-shot baseline on benchmarks such as SPair-71k, PF-PASCAL, and TSS~\citep{Min19SPair,ham2017proposal,taniai2016joint,zhang2023tale}, with noisy DINOv2 features complemented by the smoother spatial cues of diffusion models.
However, these features are learned from 2D objectives and lack explicit 3D awareness, leading to systematic failure modes~\citep{mariotti2024spherical,dunkel2025diy}.
For symmetric objects, such as cars, buses, and animals, 2D features may confuse left and right object sides \citep{Zhang:2024:Telling}.
For objects with repeated parts, such as wheels, legs, windows, or chair legs, visually similar regions may collapse to nearly identical feature representations despite corresponding to different object parts (see the nearest-neighbor visualization in \cref{fig:teaser}a).
More generally, 2D features cannot reliably distinguish structures that are visually similar yet geometrically distinct.

Several recent methods address these ambiguities by injecting a weak 3D prior to guide feature learning and correspondence filtering~\citep{mariotti2024spherical,dunkel2025diy}.
While effective, both approaches \textit{require human pose annotations and approximate object geometry with a coarse spherical proxy}, which cannot represent the geometric structure of an actual instance.
Therefore, some finer distinctions between symmetric or articulated parts are not captured.
The reliance on manual pose annotations also limits scalability, as extending to new object categories requires additional labeling effort.

In this paper, we propose a 3D-aware post-training framework that incorporates priors from 3D foundation models without requiring manual pose annotations. Given an image, we use SAM3D to estimate object geometry and pose~\citep{sam3d}, then refine the pose via a render-and-compare optimization that aligns rendered geometry with the observed object. These refined predictions allow us to run PartField~\citep{liu2025partfield} on the reconstructed shape and render geometry-aware descriptors back into the image plane, complementing DINOv2 and Stable Diffusion features in two ways. First, rendered PartField descriptors disambiguate symmetric structures and repeated parts (\eg, front vs.\ rear wheels) that 2D features alone cannot separate. Second, geodesic distances on the 3D reconstructed shape enable more reliable filtering of candidate correspondences than coarse canonical-sphere proxies, yielding higher-quality pseudo-labels for a lightweight adapter trained on top of DINOv2 and Stable Diffusion features. Experiments on standard benchmarks show consistent improvements over prior approaches with less manual supervision.
In summary, we make the following contributions:
\begin{enumerate}[label=\textit{(\roman*)}, leftmargin=*, itemsep=0.05em]
  \item a 3D-aware post-training framework for semantic correspondence that incorporates priors from 3D foundation models without human pose annotations;
  \item a render-and-compare pose refinement that allows rendering PartField features into the image plane, yielding geometry-aware features complementing DINOv2 and Stable Diffusion features;
  \item a pseudo-label filtering scheme based on geodesic distances on the estimated 3D shapes, providing higher-quality supervision than coarse spherical geometry; and
  \item geometry-aware refined features that achieve state-of-the-art semantic correspondence over prior methods with reduced manual supervision.
\end{enumerate}

\section{Related Work}
\paragraph{Semantic correspondence with foundation features}
Semantic correspondence aims to match semantically equivalent parts across object instances, which is substantially harder than low-level image matching because appearance, shape, pose, articulation, and visibility all vary. Early approaches relied on hand-crafted descriptors and learned matching networks~\citep{Lowe04SIFT,Liu11SIFTFlow,ham2017proposal,Yi16}, and because dense annotations are costly, later work explored weak supervision, cycle-consistency losses, and pseudo-label expansion from sparse labels~\citep{zhou2016learning,kim2022semi,li2021probabilistic,huang2023weakly}. Recent progress has shifted to foundation features: self-supervised vision transformers such as DINO and DINOv2 encode transferable semantic concepts~\citep{caron2021emerging,amir2022deep,oquab2023dinov2}, while text-to-image diffusion features provide complementary spatial and semantic cues~\citep{rombach2022high,hedlin2023unsupervised,tang2023emergent,luo2023diffusion,Li:2024:Sd4match}. Their fusion has become a strong zero-shot baseline~\citep{zhang2023tale}, and distillation or adapter-based refinement further improves them when supervision is available~\citep{Zhang:2024:Telling,fundel2024distillation,xue2025matcha}. However, since these features are learned from images, they remain prone to geometry-sensitive failures such as left-right confusion, front-back ambiguity, and repeated parts~\citep{Zhang:2024:Telling,mariotti2024spherical,dunkel2025diy,Mariotti:2025:Jamais}. Our work follows the weakly supervised, foundation-feature direction, but uses reconstructed 3D geometry to generate and filter dense pseudo-labels rather than relying on manual keypoint annotations.

\paragraph{Geometric priors and 3D-aware features}
A complementary line of work introduces geometric structure to disambiguate the failures of purely image-based correspondence. CAD-based cycle consistency and canonical surface mappings link image pixels to a shared object surface~\citep{zhou2016learning,canSurfMap2019abhinav,Neverova20}, while category-level templates, atlases, and learned 3D representations capture correspondences via a shared geometric frame~\citep{novum,SHIC,Common3D,semalign3d2025,chic3po}. These methods show the value of 3D structure but typically require mesh templates, precise pose, or category-level reconstruction pipelines. Closer to our setting, Spherical Maps inject a weak 3D prior by mapping image features to a category-conditioned sphere with viewpoint supervision~\citep{mariotti2024spherical}, and DIY-SC produces pseudo-labels from DINOv2 and Stable Diffusion features, then filters them against a spherical 3D prototype before training a lightweight adapter~\citep{dunkel2025diy}. In parallel, 3D foundation models make instance-level geometry practical from a single image: SAM3D reconstructs object-centric 3D shape~\citep{sam3d}, orientation models help resolve canonical-frame ambiguities~\citep{OriAny2}, and 3D feature fields or functional-map methods provide geometry-aware descriptors on surfaces~\citep{liu2025partfield,ovsjanikov2012functional,donati2020deep,dutt2024diffusion,zhu2024densematcher,wang2024gs}. In contrast to spherical-prior approaches, we combine instance-specific SAM3D meshes with PartField descriptors to both \emph{generate} and \emph{filter} pseudo-labels using faithful, per-instance 3D structure -- removing the need for manual pose annotations and coarse geometric proxies.

\section{Method}
\label{sec:method}

We estimate semantic correspondences by combining 2D foundation features with 3D geometric priors obtained from reconstructed object meshes. Our pipeline has three stages: (i) we first reconstruct and canonicalize an object-centric 3D mesh for each instance; (ii) we then render 3D-aware PartField descriptors into the image plane and use them together with DINOv2 and Stable Diffusion features to propose semantic correspondences; (iii) finally, we reject geometrically inconsistent matches using geodesic consistency on the reconstructed meshes and train a lightweight correspondence adapter on the retained pseudo-labels.

\subsection{Canonicalized 3D Object Reconstruction}
\label{sec:meth:pseudo_gt_pose}
\begin{figure}[t]
    \vspace{-3em}
    \centering
    \includegraphics[width=\linewidth]{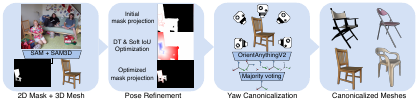}
    \caption{\textbf{Canonicalized 3D object reconstruction pipeline.} Given an image, we obtain an instance mask and a mesh from foundation models. We then refine the mesh pose via a two-phase render-and-compare optimization based on a distance-transform (DT) and a soft-IoU phase. Finally, we resolve the residual four-fold yaw ambiguity by rendering the mesh at eight known orientations and applying OrientAnything~V2 with majority voting to select the canonical yaw correction $\Delta\psi^*$.}
    \label{supp:fig:pipeline-data}
    \vspace{-1.1em}
\end{figure}

Our correspondence pipeline relies on a 3D mesh for each object instance,
expressed in a canonical frame that is consistent across instances of the same
category. We obtain such meshes from a single image without manual pose
annotation by combining recent foundation models for segmentation and
single-image 3D reconstruction with two refinement stages.
While these foundation models provide a strong geometric prior, their outputs
exhibit two systematic issues: the predicted scale and translation can be
inaccurate, causing the rendered mesh to misalign with the image, and the
canonical orientation is ambiguous up to discrete yaw rotations across
instances. We address the first issue with a render-and-compare optimization
that aligns the rendered silhouette to the observed mask, and the second with a yaw canonicalization
step based on multi-view orientation estimation. The full process is illustrated in
\cref{supp:fig:pipeline-data}.

\paragraph{2D Mask and 3D Mesh Initialization}
We extract a 2D instance mask $\mathbf{M} \in \{0,1\}^{H \times W}$ with
SAM3~\citep{sam3}, using the image together with the dataset-provided category label. Given this mask, SAM3D~\citep{sam3d} reconstructs an
object-centric mesh from the masked image in a feed-forward manner and
additionally predicts the camera parameters used for rendering. In the following, we show how we refine and canonicalize this initial reconstruction.

\paragraph{Render-and-Compare Pose Refinement}
To correct the residual scale and translation error in the SAM3D
reconstruction, we apply a render-and-compare optimization on top of the
predicted camera. Concretely, we optimize a scale factor
$s = e^{\ell} \in \mathbb{R}_{>0}$ (parameterized in log-space to remain
strictly positive) and a translation $\mathbf{t} \in \mathbb{R}^{3}$ applied to
the mesh, by minimizing the discrepancy between the rendered
soft silhouette $\hat{\mathbf{M}}(s, \mathbf{t}) \in [0,1]^{H \times W}$ and
the observed mask $\mathbf{M}$. Since the soft IoU between $\hat{\mathbf{M}}$
and $\mathbf{M}$ has no gradient when the two are disjoint, we proceed in two
sequential phases: a distance-transform (DT) phase that provides a global gradient
signal regardless of initial alignment, followed by a soft-IoU phase that
sharpens the fit.

\paragraphit{Distance-transform attraction}
We first dilate $\mathbf{M}$ by $r$ to obtain $\tilde{\mathbf{M}}$, providing tolerance for coarse mesh boundaries, and compute two squared distance fields normalized by the image diagonal $d$:
\begin{equation}
  \mathcal{D}_{\text{out}}(p) = \tfrac{1}{d}
      \min_{p':\,\tilde{\mathbf{M}}(p')=1} \|p - p'\|_{2}^{2},
  \qquad
  \mathcal{D}_{\text{in}}(p) = \tfrac{1}{d}
      \min_{p':\,\tilde{\mathbf{M}}(p')=0} \|p - p'\|_{2}^{2}.
  \label{eq:dist_transform}
\end{equation}
$\mathcal{D}_{\text{out}}$ is zero inside $\tilde{\mathbf{M}}$ and grows with
distance to the mask; $\mathcal{D}_{\text{in}}$ is zero outside
$\tilde{\mathbf{M}}$ and grows with depth into its interior. The DT loss
combines these into a mask-alignment objective:
\begin{equation}
  \mathcal{L}_{\text{DT}}
    = \frac{1}{HW}\sum\nolimits_{p}
      \Bigl[
        \hat{\mathbf{M}}_{p}\,\mathcal{D}_{\text{out}}(p)
        +
        \mathcal{D}_{\text{in}}(p)
          \bigl(1 - \lambda\,\hat{\mathbf{M}}_{p}\bigr)
      \Bigr].
  \label{eq:dt_loss}
\end{equation}
The first term pulls rendered mass that falls outside the mask back toward it,
weighted by how far outside it is. The second term simultaneously penalizes
uncovered mask interior \emph{and}, through the coefficient $\lambda > 1$,
rewards rendered coverage of the interior. Without this reward, the
optimization tends to under-cover the mask under partial occlusion — the
rendered silhouette settles on a small fully-contained region rather than
extending to the occluded extent of the object. 

\paragraphit{Soft-IoU refinement}
Once the rendered and observed masks overlap, the soft IoU has a usable
gradient and we switch to a differentiable soft-IoU loss:
\begin{equation}
  \mathcal{L}_{\text{IoU}}
    = 1 -
      \frac{\sum_{p} \hat{\mathbf{M}}_{p}\,\mathbf{M}_{p}}
           {\sum_{p}\bigl(\hat{\mathbf{M}}_{p} + \mathbf{M}_{p}
                          - \hat{\mathbf{M}}_{p}\,\mathbf{M}_{p}\bigr)}.
  \label{eq:iou_loss}
\end{equation}
This phase tightens the alignment that the previous phase has approximately
established. 

\paragraph{Yaw Canonicalization}
Even after pose refinement, SAM3D meshes do not necessarily share a
consistent canonical orientation across instances of the same category. We
find that roughly $6\%$ of meshes are misaligned by a multiple of $90^{\circ}$ around the vertical axis — a four-fold yaw ambiguity that is most common for symmetric or elongated objects such as buses, boats, and trains.
To resolve this without manual annotation, we use OrientAnything
V2~\citep{OriAny2} as an external orientation estimator. For each mesh, we
render eight views at known yaw angles
$\psi_{\text{known}} \in \{0^{\circ}, 45^{\circ}, \ldots, 315^{\circ}\}$ and
we estimate the apparent orientation $\psi_{\text{est}}$
of each rendering. If the mesh is correctly canonicalized, $\psi_{\text{est}}$
should match $\psi_{\text{known}}$ up to estimator noise; otherwise, the two
differ by a multiple of $90^{\circ}$. For each rendered view we therefore pick
the discrete correction that best closes this gap,
\begin{equation}
    \Delta\psi^* =
    \underset{\Delta\psi \,\in\,
      \{0^{\circ}, 90^{\circ}, 180^{\circ}, 270^{\circ}\}}{\arg\min}
    \bigl|\psi_{\text{est}} + \Delta\psi - \psi_{\text{known}}\bigr|,
    \label{eq:canon}
\end{equation}
and aggregate the eight candidates into a single one by majority vote, which makes the procedure robust to occasional orientation estimation errors. Each mesh is then rotated by the selected $\Delta\psi^*$, yielding a set of consistently canonicalized meshes that serve as the geometric backbone for what follows. 

\begin{figure}
    \vspace{-2em}
    \centering
    \includegraphics[width=\linewidth]{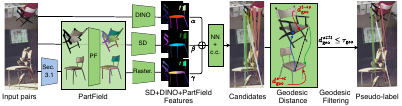}
    \caption{\textbf{Pseudo-label correspondences pipeline.} Given two images, we fuse DINO, SD, and PartField features (rasterized from the meshes of \cref{sec:meth:pseudo_gt_pose}) and propose candidate matches via nearest-neighbor (NN) search with relaxed cyclic consistency (c.c). Each candidate is then geometrically verified by lifting the matched pixels onto the reconstructed meshes and computing the geodesic error $d_\text{geo}^{s \rightleftarrows t}$; candidates exceeding threshold $\tau_\text{geo}$ are rejected. The retained pseudo-labels $\mathcal{P}$ are used to train a lightweight correspondence adapter on top of frozen DINO+SD features.}
    \label{fig:pipeline-corresp}
    \vspace{-1.1em}
\end{figure}
\subsection{Pseudo-Label Semantic Correspondences}\label{sec:meth:pseudo_gt_corresp}
Given a pair of images of the same object category, we generate correspondence pseudo-labels in two stages. First, we fuse 2D foundation features (DINO+SD) with 3D-aware PartField features rasterized from the canonicalized meshes, and apply relaxed cyclic consistency to discard obvious mismatches. Second, each surviving candidate is verified geometrically: matched points are lifted onto their respective meshes and rejected if their geodesic distance exceeds a threshold. The two stages are complementary --- cyclic consistency is a cheap image-space filter, while geodesic verification is a geometry-grounded confidence measure that exploits the 3D shapes from \cref{sec:meth:pseudo_gt_pose}.\\
\paragraphit{Notation}
We use the superscripts $\square^s$ and $\square^t$ to denote quantities
associated with the source and target, respectively. We denote $\PointPixel$ as a point in image space while $\PointThreeD$
denotes a point in the 3D space.

\paragraph{PartField Features}
PartField~\citep{liu2025partfield} (PF) predicts a continuous per-vertex feature field encoding geometric and part-level structure directly from the 3D shape $\Shape$. These descriptors naturally distinguish parts that are visually similar but geometrically distinct (\eg, front vs.\ rear wheels, left vs.\ right legs), exactly the cases where 2D foundation features tend to collapse. 
To use PartField in image space, we rasterize the per-vertex descriptors
into the input image using the SAM3D camera together with the refined pose from \cref{sec:meth:pseudo_gt_pose}. Vertices outside the camera frustum
or outside the foreground mask are discarded, and foreground pixels with no
projected descriptor are filled by nearest-neighbor propagation. The result
is an image-space PartField map aligned with the RGB image, which can be
combined with 2D image features for semantic correspondence estimation.  PCA visualizations and rasterization details are deferred to Supp. \cref{supp:feature-partfield}.

\paragraph{Candidate Generation}
Given the fused image-space DINO+SD+PF features, we propose candidate matches via nearest-neighbor search, retaining only those that pass a cyclic consistency check.

\paragraphit{Feature fusion}
Following \citet{zhang2023tale}, we fuse our three feature sources by
independently L2-normalizing each and concatenating them with category-agnostic
weights. We denote the normalized feature vectors as 
$
\widehat{\Feature}_{s} = \frac{\Feature_{s}}{\|\Feature_{s}\|_2}.
$
The fused representation is then defined as
\begin{equation}\label{eq:fused_features}
\Feature_{\text{fused}} =
\bigl(
\sqrt{\alpha}\,\widehat{\Feature}_{\text{SD}},\;\;
\sqrt{\beta}\,\widehat{\Feature}_{\text{DINO}},\;\;
\sqrt{\gamma}\,\widehat{\Feature}_{\text{PF}}
\bigr),\quad \text{with } \gamma = 1 - \alpha - \beta.
\end{equation}
We use the weights $\alpha =1/2$, $\beta = 1/3$, and $\gamma =1/6$, which we found offers a good
balance between the three features in practice; a weight sweep is provided in
Supp. \cref{supp:feature-weights-search}. Candidate matches are then
proposed by nearest-neighbor search in the fused space.

\paragraphit{Relaxed cyclic consistency}
While the 3D-aware PartField features significantly enhance the matching quality, some candidates remain wrongly matched.
To filter these mismatches, we apply a relaxed cyclic consistency
check inspired by \citet{aberman2018neural}. 
As observed in \citet{dunkel2025diy}, strict cyclic consistency rejects a large fraction of correct matches due to sub-pixel noise; we therefore relax the criterion to require only that the backward match lies within a small spatial tolerance of the source. A candidate $(\PointPixel^s, \hat{\PointPixel}^t)$ with
$\hat{\PointPixel}^t = \text{NN}_\text{fused}^{s \rightarrow t}(\PointPixel^s)$
is retained if
\begin{equation}
    \bigl\| \text{NN}_\text{fused}^{t \rightarrow s}(\hat{\PointPixel}^t) - \PointPixel^s \bigr\|_2
    < \tau_{cc} \cdot \max(h, w),
\end{equation}
where $\text{NN}_\text{fused}$ denotes nearest-neighbor search in the fused
feature space, $h$ and $w$ are the object's bounding
dimensions, and $\tau_{cc}$ is a tolerance ratio.\\
\paragraph{Candidate Verification via Geodesic Filtering}
\label{sec:meth:pseudo_gt_corresp:geodesic}
Our fused descriptor uses a fixed mixing strategy, and these fused features inevitably produce some wrong matches since objects greatly vary across instances. Cyclic consistency removes some of them but operates purely in feature space, ignoring 3D geometry. We therefore add a geodesic consistency stage: matched locations lifted onto canonically posed meshes must land in nearby surface regions.\\

\paragraphit{Lifting matches to 3D}
Given a candidate match $(\PointPixel^s, \PointPixel^t)$, we cast a ray from each camera through the corresponding pixel and intersect it with the respective mesh, obtaining the unprojected points $\PointThreeD^s$ and $\PointThreeD^t$ together with their containing triangles and barycentric coordinates. Because geodesic distances are computed between mesh vertices, we snap each unprojected point to the dominant vertex of its triangle (the vertex with the largest barycentric weight), giving $\bar{\PointThreeD}^s$ and $\bar{\PointThreeD}^t$.\\

\paragraphit{Cross-mesh correspondence via PartField}
The previous step places each candidate match onto the source and target
meshes individually. However, the meshes share only a canonical orientation but not vertex correspondence. To compare the lifted source and target points, we therefore estimate a 3D correspondence between the meshes themselves. Hence, we use PartField nearest-neighbor as the cross-mesh correspondence: we interpolate the PartField descriptor at $\PointThreeD^s$ on the source mesh and search
for its nearest neighbor among the PartField descriptors on the target mesh,
$
    \hat{\PointThreeD}^t = \text{NN}_\text{PF}^{s \rightarrow t}(\PointThreeD^s),
$
yielding a target vertex $\hat{\PointThreeD}^t$ that represents the cross-mesh counterpart of $\PointThreeD^s$. 
A candidate is then geometrically consistent if this PartField-predicted target $\hat{\PointThreeD}^t$
is geodesically close to the target obtained from the image-space match,
$\bar{\PointThreeD}^t$.\\

\paragraphit{Bicyclic geodesic error}
We measure the disagreement between the two target predictions as a
\emph{bicyclic} geodesic distance, combining a forward and a backward
geodesic error on the source and target meshes. The forward error measures,
on the target mesh, the geodesic distance between the cross-mesh prediction
$\hat{\PointThreeD}^t$ and the target $\bar{\PointThreeD}^t$ obtained from
the image-space match:
\begin{equation}
    d_\text{geo}^{s \rightarrow t}
    =
    d_{\Mesh_t}\bigl(\hat{\PointThreeD}^t,\, \bar{\PointThreeD}^t\bigr).
\end{equation}
A symmetric computation in the reverse direction yields a backward error
$d_\text{geo}^{t \rightarrow s} = d_{\Mesh_s}(\hat{\PointThreeD}^s,\, \bar{\PointThreeD}^s)$,
where $\hat{\PointThreeD}^s = \text{NN}_\text{PF}^{t \rightarrow s}(\PointThreeD^t)$.
We average the two and normalize by the mesh bounding-box diagonals so that
the score is comparable across instances and categories of varying scale:
\begin{equation}
    d_\text{geo}^{s \rightleftarrows t}
    =
    \frac{1}{2}
    \left(
    \frac{d_\text{geo}^{s \rightarrow t}}{\mathrm{diag}(\Mesh_t)}
    +
    \frac{d_\text{geo}^{t \rightarrow s}}{\mathrm{diag}(\Mesh_s)}
    \right).
\end{equation}
Intuitively, $d_\text{geo}^{s \rightleftarrows t}$ is small when the
image-space candidate and the PartField cross-mesh correspondence agree on
the same surface location, and large when they disagree.

\paragraphit{Rejection of wrong pseudo-labels}
We use the bicyclic geodesic error as a per-candidate quality score and
threshold it to reject inconsistent pseudo-labels. A candidate
$(\PointPixel^s, \PointPixel^t)$ is retained if and only if its error falls below a threshold $\tau_\text{geo}$:
\begin{equation}
    \mathcal{P}
    =
    \left\{
    (\PointPixel^s, \PointPixel^t)
    \;\middle|\;
    d_\text{geo}^{s \rightleftarrows t} \leq \tau_\text{geo}
    \right\}.
\end{equation}
Because $d_\text{geo}^{s \rightleftarrows t}$ is normalized by the mesh
bounding-box diagonals, a single value of $\tau_\text{geo}$ applies across object instances and categories of varying scale.
Crucially, we do not require correspondences to cover every object part: obtaining fewer but geometrically reliable pseudo-labels is preferable to dense but noisy supervision, since the adapter only benefits from matches it can trust.

\paragraph{Supervised Training with Pseudo-Labels}

We use the pseudo-labels $\mathcal{P}$ to train a lightweight
adapter $f_p(\cdot)$ on top of frozen DINOv2 and Stable Diffusion features,
following \citet{dunkel2025diy}. The adapter has been shown to outperform
zero-shot feature concatenation~\citep{zhang2023tale, Zhang:2024:Telling} and
weighted feature combinations with weak geometric
regularization~\citep{mariotti2024spherical}, while keeping the underlying
foundation features unchanged. We denote the adapted features by
$\Feature^s$ and $\Feature^t$ for the source and target images, respectively.
We supervise $f_p(\cdot)$ with two complementary losses. A sparse contrastive
loss~\citep{luo2023diffusion} acts on the labeled pseudo-correspondences,
maximizing similarity between matched points and minimizing it against
non-matching points:
\begin{equation}
    \mathcal{L}_{\text{sparse}}
    = \mathrm{CL}\bigl(\Feature^s(\SetPixelPoint^s),\,
                       \Feature^t(\SetPixelPoint^t)\bigr).
\end{equation}
A dense regression loss~\citep{Zhang:2024:Telling} additionally propagates
gradients to image regions without explicit labels by predicting the target
location with a window soft-argmax over the feature similarity map and
penalizing its deviation from the labeled target:
\begin{equation}
    \mathcal{L}_{\text{dense}}
    = \sum_{(\PointPixel^s, \PointPixel^t) \in \mathcal{P}}
      \bigl\| \hat{\PointPixel}^t - (\PointPixel^t + \epsilon) \bigr\|_2,
    \qquad
    \hat{\PointPixel}^t = \textsc{WindowSoftArgmax}\!\bigl(
        \Feature^s(\PointPixel^s)^\top \Feature^t \bigr),
\end{equation}
where $\epsilon$ is small Gaussian noise that regularizes the predicted
location at sub-pixel scale. The adapter is trained with the sum
$\mathcal{L} = \mathcal{L}_{\text{sparse}} + \mathcal{L}_{\text{dense}}$.

\section{Experiments}
\label{sec:exp}
\begin{table*}[t]
\vspace{-2em}
\centering
\caption{\textbf{Evaluation on standard benchmarks.} 
Per-image PCK (\%, $\uparrow$) at multiple thresholds on SPair-71k (test and Geo-Aware subset), AP-10K and SpairU. `\text{--}' indicates missing numbers. Best per method type is shown in \textbf{bold}. Full Table including \textit{Supervised} methods can be found in \cref{supp:tab:full_table}.}
\label{tab:benchmark_spair_ap}
\setlength{\tabcolsep}{4pt}
\footnotesize
\begin{tabularx}{\linewidth}{@{}Xr|ccc|ccc|ccc|ccc@{}}
\toprule
 && \multicolumn{3}{c}{\textbf{SPair-71k}} 
 & \multicolumn{3}{c}{\textbf{SPair-Geo-Aware}} 
& \multicolumn{3}{c}{\textbf{AP-10K} (0.10)} 
& \multicolumn{3}{c@{}}{\textbf{SpairU}} \\
\cmidrule(lr){3-5}\cmidrule(lr){6-8}\cmidrule(lr){9-11}\cmidrule(lr){12-14}
\textbf{Method} &&
0.01 & 0.05 & 0.10 &
0.01 & 0.05 & 0.10 
& I.S. 
& C.S.  
& C.F.
& 0.01 & 0.05 & 0.10  \\
\midrule
\multicolumn{11}{@{}l}{\textit{Unsupervised}} \\
\myrowcolour
~~DINOv2+NN&\citet{zhang2023tale}
& 6.3 & 38.4 & \textbf{53.9} 
& \textbf{3.4} & \textbf{28.2} & 42.0
& \textbf{60.9}
& \textbf{57.3}
& \textbf{47.4} 
& \text{--} & \text{--} & \textbf{54.9} \\
~~DIFT&\citet{tang2023emergent}
& \textbf{7.2} & \textbf{39.7} & 52.9 
& \textbf{3.4} & \textbf{28.2} & \textbf{42.5}
& 50.3 
& 46.0 
& 35.0 
& \text{--} & \text{--} & 47.4 \\
\midrule
\multicolumn{11}{@{}l}{\textit{Weakly supervised with human annotations}} \\
\myrowcolour
~~Spherical Map.& \citet{mariotti2024spherical}
& \text{8.4} & \text{48.2} & \text{64.4}
& \text{--} & \text{--} & \text{--}
& 65.4 & 63.1 & 51.0
& \text{--} & \text{--} & 61.0\\
~~DIY-SC& \citet{dunkel2025diy}
& \textbf{10.1} & \textbf{53.8} & \textbf{71.6} 
& 7.7 & 47.7 & 67.5
& \textbf{70.6}
& \textbf{69.8}
& \textbf{57.8} 
& 5.4 & 44.0 & \textbf{67.9} \\

\midrule
\multicolumn{11}{@{}l}{\textit{Weakly supervised without human annotations}} \\

\myrowcolour
~~SD{+}DINOv2&\citet{zhang2023tale}
& 7.9 & 44.7 & 59.9 
& 5.3 & 34.5 & 49.3
& 62.9 
& 59.3 
& 48.3 
& \text{--} & \text{--} & 59.4\\
~~DIY-SC+OriAny&\citet{dunkel2025diy} 
& 9.5 & 51.2 & 69.6
& 6.9 & 45.7 & 65.8
& 69.3
& 66.8
& 54.0
& 5.2 & 43.1 & 66.3 \\
\myrowcolour
~~\textbf{\ours (Ours)}& 
& \textbf{10.2} & \textbf{54.8} & \textbf{73.0}
& \textbf{7.8} & \textbf{50.1} & \textbf{70.8}
& \textbf{69.6}
& \textbf{68.5}
& \textbf{56.9}
& \textbf{5.6} & \textbf{43.5} & \textbf{67.3} \\
\bottomrule
\end{tabularx}
\vspace{-1em}
\end{table*}

In this section, we evaluate on four semantic correspondence benchmarks against unsupervised and weakly supervised baselines, and ablate the key components of our pipeline.

\subsection{Implementation Details}
\label{sec:exp:implementation}

We report the values for the method parameters introduced in \cref{sec:method}.
For the distance-transform objective in \cref{eq:dt_loss}, we set the interior-coverage reward to $\lambda=4$. During pose refinement, we optimize log-scale and translation with Adam using separate learning rates $\text{lr}_\text{scale}=0.05$ and $\text{lr}_\text{trans}=0.02$.
Similarly to \citet{zhang2023tale}, we extract SD and DINO features from images resized to $960^2$ (SD, DINOv3) and $840^2$ (DINOv2). PartField descriptors are rasterized at the shared correspondence-map resolution of $60^2$.  For the relaxed cyclic consistency, we set the tolerance to $\tau_{cc}=0.05$ of the object's bounding box, with a lower bound of one feature-map patch. For geometric verification, we use $\tau_\text{geo}=0.05$.
Following prior work~\citep{Zhang:2024:Telling,dunkel2025diy}, we use a four-layer, 5M-parameter adapter. We train it with AdamW with $\text{lr}=5{\cdot}10^{-3}$, weight decay of $10^{-3}$, and a one-cycle schedule for 200k iterations. Each image pair has $\sim\!1600$ pseudo-annotations; we sample 50 per iteration to prevent denser pairs from dominating training. More details in Supp. \cref{supp:sec:implementation_details}.

\paragraph{Benchmarks and Metrics}
We evaluate on four standard semantic correspondence benchmarks.
\textit{SPair-71k}~\cite{Min19SPair} contains 71k image pairs across 18 categories, with up to 20 keypoints per image and up to 900 images per category. Following \citet{Zhang:2024:Telling}, we additionally report results on \textit{SPair-Geo-Aware}, a subset of SPair-71k that emphasizes challenging correspondences involving symmetric or repeated parts and therefore better tests whether a method correctly captures object orientation and geometry.
\textit{SPair-U}~\cite{Mariotti:2025:Jamais} extends SPair-71k with $\sim4$ additional unseen keypoints per category, providing an interesting evaluation of keypoint-level generalization.
\textit{AP-10K}~\cite{Yu:2021:AP10k} is an animal pose dataset with 17 keypoints shared across 54 species spanning intra-species, cross-species, and cross-family matching.
Following prior work~\citep{zhang2023tale}, we use the Percentage of Correct Keypoints (PCK@$\alpha$) as metric, for which a prediction is considered correct if it lies within a distance of \(\alpha \cdot \max(h, w)\) from the ground-truth keypoint with \(h, w\) the object's bounding-box dimensions. We only report the most common metric: per-image PCK averaged over the test set.

\paragraph{Baselines}
We compare our performances with recent works which we categorize into 4 different categories: \textit{Unsupervised}, \textit{Weakly Supervised with Human Annotations required}, and \textit{Weakly Supervised without Human Annotations} (\ours's category). Our focus will remain on the \textit{Unsupervised} and \textit{Weakly supervised} approaches. DIFT~\citep{luo2023diffusion}, and SD + DINOv2 \citep{zhang2023tale} extract features from foundation models and perform nearest-neighbor matching in feature space. Spherical mapper~\citep{mariotti2024spherical} and DIY-SC~\citep{dunkel2025diy} both leverage pose annotations as weak supervision during training.

\subsection{Experimental Results}
\paragraph{Evaluation on SPair-71k}
As shown in \cref{tab:benchmark_spair_ap}, \ours{} establishes the strongest results among weakly supervised methods on SPair-71k, reaching 73.0 \pckTen. In particular, it improves over the strongest baseline in the same supervision regime, DIY-SC+OriAny, by 3.4 points. Per-category results in \cref{supp:tab:Spair_all_cats} show that the gains are concentrated in rigid categories with strong geometric symmetry, such as bus ($+10.8$), tv monitor ($+9.8$), car ($+6.9$), and motorcycle ($+5.1$), while non-rigid categories such as animals show no gain or can slightly regress.
The gains are even more pronounced on \textit{SPair-Geo-Aware}, where our method reaches 70.8 \pckTen, clearly surpassing all existing weakly supervised approaches. This behavior is consistent with our central hypothesis: because our pseudo-labels are grounded in reconstructed 3D geometry, they are especially effective on correspondences that require disambiguating symmetric or repeated parts, and viewpoint changes.

\paragraph{Evaluation on SPairU}
On SPairU, \ours{} obtains 67.3 \pckTen. This is the best result among methods without human annotations and is only 0.6 points below DIY-SC, which leverages human annotations. 
The smaller margin compared with SPair-Geo-Aware subset is expected: SPairU mainly probes generalization to previously unseen keypoints which are usually located at the middle of the limbs/parts. Our PartField features, trained on part contrastive learning, are not explicitly designed to differentiate keypoints within the same part. Hence we do not expect a large gain from PartField features on this benchmark, explaining the modest improvement over DIY-SC+OriAny (1 point).
Nevertheless, the result shows that the representation learned from our pseudo-labels transfers also to these keypoint definitions.

\paragraph{Evaluation on AP-10K}
Our method also transfers well to the more articulated and shape-diverse setting of AP-10K. \ours achieves 69.6/68.5/56.9 \pckTen{} on the intra-species, cross-species, and cross-family splits, outperforming the strongest baseline without human annotations on all three splits. These improvements are particularly meaningful on the harder cross-species and cross-family evaluations, where appearance cues alone are often insufficient. Although PartField descriptors can be less reliable for animals in unusual poses during the pseudo-annotation procedure, the overall results show that our 3D-aware pseudo-label generation and filtering pipeline remains effective well beyond the rigid object categories of SPair-71k.

\begin{figure}
\vspace{-2em}
    \centering
    \includegraphics[width=\linewidth]{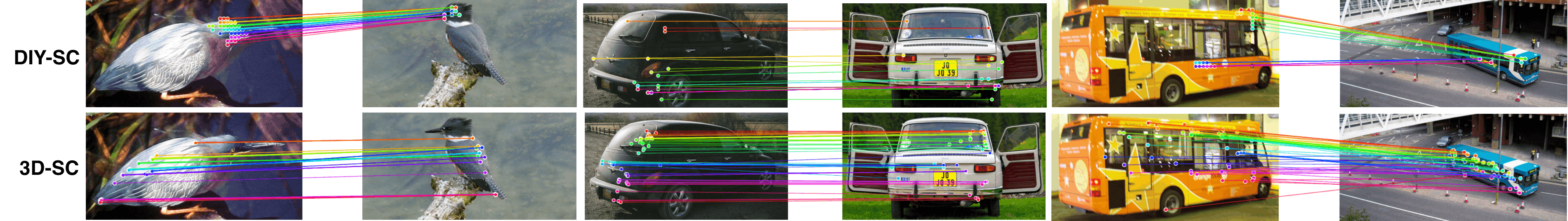}
    \caption{\textbf{Qualitative pseudo-annotations.} We visualize pseudo-ground-truth annotations from \ours and DIY-SC. \ours{} produces denser and more geometrically consistent pseudo-annotations.}
    \label{fig:pseudo-annot-quali}
    \vspace{-1.1em}
\end{figure}

\paragraph{Qualitative results}
As shown in \cref{fig:pseudo-annot-quali} (additional visualizations in \cref{supp:fig:pseudo-annot-quali}), \ours{} produces well-distributed pseudo-annotations that cover the commonly visible parts of each object. The matches are geometrically consistent and free from left-right ambiguities, a direct consequence of anchoring correspondence in instance-specific 3D geometry.

\subsection{Ablations}
\begin{table}
\vspace{-2em}
\centering

\begin{minipage}{0.42\linewidth}
\centering
\caption{\textbf{Filtering evaluation on validation set.} FPR refers to False Positive Rate, \ie, unfiltered wrong prediction. \#Candidates refers to the average total number of filtered correspondences per pair.}
\label{tab:filter_validation}
\vspace{0.4em}
\footnotesize
\begin{tabularx}{\linewidth}{X@{\hskip 0.1cm}|c@{\hskip 0.1cm}|c}
\toprule
Filter & FPR & \#Candidates \\
\midrule
\multicolumn{3}{@{}l}{Features: \textit{SD+DINO}} \\
\myrowcolour
~~Spherical mapper
       &     10.95        &   1856    \\
~~Triplane
       &    13.15        &   1948    \\
\myrowcolour
~~PF Feature similarity
       &       2.81     &   1608    \\
~~Geodesic
       &      1.82     &   1543    \\
\midrule
\multicolumn{3}{@{}l}{Features: \textit{SD+DINO+PartField}} \\
\myrowcolour
~~Spherical mapper
       &      10.75       &  2001     \\
~~Triplane
       &      13.07      &  2090     \\
\myrowcolour
~~PF Feature similarity
       &     2.47       &  1694     \\
~~Geodesic
       &     1.78      &   1634    \\
\bottomrule
\end{tabularx}
\end{minipage}
\hfill
\begin{minipage}{0.54\linewidth}
\caption{\textbf{Ablations on SPair-71k.} All introduced components bring a significant improvement. The baseline is evaluated using the SD+DINO zero-shot approach with window soft argmax. `c.c.' = cyclic consistency.}
\label{tab:ablations}
\footnotesize
\centering
\begin{tabularx}{\linewidth}{
>{\centering\arraybackslash}X@{\hskip 0.1cm}
|>{\centering\arraybackslash}X@{\hskip 0.1cm}
|>{\centering\arraybackslash}X@{\hskip 0.1cm}
|>{\centering\arraybackslash}X@{\hskip 0.1cm}
|>{\centering\arraybackslash}X@{\hskip 0.1cm}
|>{\centering\arraybackslash}X@{\hskip 0.1cm}
|>{\centering\arraybackslash}X}
\toprule
pseudo & PF & c.c. & filter. & sampl. & DINO & PCK@.1 \\
\midrule
\myrowcolour
            &           &              
            &           & \checkmark
            &     v2      &    64.9  \\
            
\checkmark  &            &              
            &          & \checkmark
            &     v2      &   67.0    \\
            
\myrowcolour
\checkmark  &            &     \checkmark        
            &          & \checkmark
            &     v2      &    67.6   \\

\checkmark  &           &    
\checkmark  &  \checkmark      & \checkmark
            &     v2      &   71.6    \\
\myrowcolour
\checkmark   &           &       
\checkmark   &     \checkmark     & \checkmark
             &     v3      &   72.4    \\
             
\checkmark   &    \checkmark        &              
            &          & \checkmark
            &     v2      &    66.9   \\
            
\myrowcolour
\checkmark   &    \checkmark        &  
\checkmark  &          & \checkmark
            &     v2      &    68.8   \\
            
\checkmark  &     \checkmark       &    \checkmark          
            &    \checkmark      & \checkmark
            &      v2     &    72.1   \\
            
\myrowcolour
\checkmark  &     \checkmark       &    \checkmark          
            &    \checkmark      & 
            &     v3     &    72.4   \\
            
\checkmark  &    \checkmark        &       \checkmark       
            &   \checkmark       & \checkmark
            &      v3     &   \textbf{73.0}    \\
\hline
\myrowcolour
    \multicolumn{5}{l|}{~~DIY-SC} &  v3 & 72.1\\
    \multicolumn{5}{l|}{~~DIY-SC+OriAny} &  v3 & 70.4\\
\bottomrule

\end{tabularx}
\end{minipage}
\vspace{-1.1em}
\end{table}

\paragraph{PartField Features}
\cref{tab:filter_validation} reports the effect of adding PartField to the feature fusion.
Compared to SD+DINO alone, SD+DINO+PartField simultaneously lowers the False Positive Rate (FPR) of unfiltered candidates and increases the average number of candidates retained per pair.
These two effects together indicate that integrating PartField not only suppresses incorrect matches but also surfaces additional correct ones that SD+DINO misses, consistent with its ability to distinguish geometrically distinct but visually similar regions such as front and rear wheels or left and right parts.
The downstream impact is confirmed in \cref{tab:ablations}: adding PartField improves \pckTen{} on SPair-71k by 0.6 points over the SD+DINO baseline.

\paragraph{Filtering}
We validate the geodesic filtering stage on the SPair-71k validation set.
For each annotated keypoint we compute its nearest neighbor in the fused feature space and check whether the prediction is correct under \pckTen.
A wrong prediction that survives filtering counts as a false positive; the FPR is therefore the fraction of unfiltered predictions that are incorrect.
As shown in \cref{tab:filter_validation}, our bicyclic geodesic filter achieves the lowest FPR of 1.78\% among the filtering strategies we compared.
The benefit also carries over to the trained adapter: \cref{tab:ablations} shows a gain of 3.3 \pckTen{} points when geodesic filtering is applied versus using all cyclic-consistency candidates without further rejection. Finally, capping the number of pseudo-labels sampled per pair during training improves \pckTen{} by 0.6 points; without this cap, pairs with denser pseudo-label sets dominate the gradient and reduce the effective diversity of training signal.

\paragraph{Backbone}
Replacing DINOv2 with DINOv3 as the vision backbone yields an improvement of 0.9 \pckTen{}.
To disentangle this gain from other design choices, we applied the same substitution to both DIY-SC variants and observe a similarly sized increase, confirming that a slight improvement (0.5-0.9) is attributable to the stronger backbone in general.
Importantly, our method outperforms both DIY-SC variants in either backbone setting.

\section{Limitations and Future Work}\label{sec:limitation}
Our pipeline depends on SAM3D's pose and shape estimates; errors propagate through the 2D--3D reprojection and can degrade geodesic consistency, although our filtering removes most resulting false positives. PartField's part-level contrastive training provides coarse regional cues rather than precise within-part localization, which motivates its relatively low fusion weight; this limitation is reflected in our SPairU results, where keypoints often lie in the middle of parts and PartField contributes less signal. A stronger 3D feature, ideally one tailored to deformable categories such as animals, would likely warrant a higher weight. Finally, our cross-mesh correspondence uses nearest-neighbor matching in PartField space; replacing it with denser registration via optimal transport or functional maps~\citep{ovsjanikov2012functional} is a natural next step, trading additional compute for finer alignment.
\section{Conclusion}
\label{sec:conclusion}
We presented a 3D-aware post-training framework for semantic correspondence that leverages priors from 3D foundation models without requiring human pose annotations. By combining SAM3D-based geometry and pose estimation with a render-and-compare refinement step, we obtain instance-specific 3D structure that drives both feature construction and pseudo-label filtering: PartField descriptors rendered into the image plane provide geometry-aware cues that complement DINO and Stable Diffusion features, while geodesic distances on the reconstructed shapes enable principled filtering of inconsistent matches. The filtered correspondences supervise a lightweight adapter that yields consistent improvements over prior methods on standard benchmarks. Our results suggest that instance-specific 3D structure can be a more reliable geometric prior than the coarse spherical proxies used by prior post-training approaches, and that it can be obtained automatically from off-the-shelf 3D foundation models. 
We see this as an early step toward a new class of self-supervised pipelines where 3D foundation models act as geometric teachers for 2D tasks, a direction that becomes more powerful as 3D reconstruction quality continues to improve.

\begin{ack}
AK acknowledges support via his Emmy Noether Research Group funded by the German Research Foundation (DFG) under grant number 468670075.
This research was funded by the Deutsche Forschungsgemeinschaft (DFG, German Research Foundation) under grant number 539134284, through EFRE (FEIH\_2698644) and the state of Baden-Württemberg. 
\begin{center}
\includegraphics[width=0.3\textwidth]{figures/acknowledgement/BaWue_Logo_Standard_rgb_pos.png} ~~~ \includegraphics[width=0.3\textwidth]{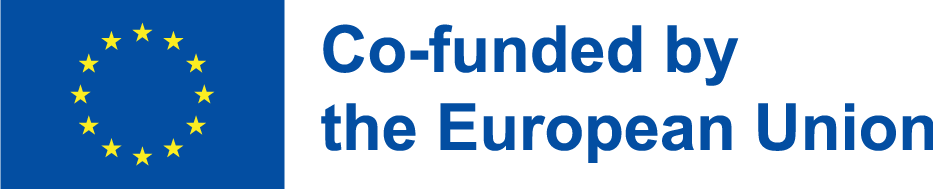} 
\end{center}
\end{ack}
{
    \bibliographystyle{abbrvnat}
    \bibliography{main}

@String(IJCV = {International Journal of Computer Vision (IJCV)})

@String(CVPR= {Proceedings of the IEEE Conference on Computer Vision and Pattern Recognition (CVPR)})

@String(ICCV= {Proceedings of the IEEE/CVF International Conference on Computer Vision (ICCV)})

@String(WACV= {Proceedings of the Winter Conference on Applications of Computer Vision (WACV)})

@String(ECCV= {European Conference on Computer Vision (ECCV)})

@String(ECCVW= {European Conference on Computer Vision Workshop (ECCVW)})

@String(NIPS= {Conference on Neural Information Processing Systems (NeurIPS)})

@String(ICLR = {International Conference on Learning Representations (ICLR)})

@String(SIGGRAPHAsia = {ACM Trans. Graphics (Proc. SIGGRAPH Asia)})

@String(PAMI = {IEEE Transactions on Pattern Analysis and Machine Intelligence (PAMI)})

@String(ThreeDV = {International Conference on 3D Vision (3DV)})

@String(TOG = {ACM Transactions on Graphics (TOG)})

@inproceedings{caron2021emerging,
  title     = {Emerging Properties in Self-Supervised Vision Transformers},
  author    = {Caron, Mathilde and Touvron, Hugo and Misra, Ishan and J{\'e}gou, Herv{\'e} and Mairal, Julien and Bojanowski, Piotr and Joulin, Armand},
  booktitle = ICCV,
  pages     = {9650--9660},
  year      = {2021}
}

@inproceedings{amir2022deep,
  title     = {Deep {ViT} Features as Dense Visual Descriptors},
  author    = {Amir, Shir and Gandelsman, Yossi and Bagon, Shai and Dekel, Tali},
  booktitle = ECCVW,
  year      = {2022}
}

@inproceedings{cuttano2026marco,
  title     = {{MARCO}: Navigating the Unseen Space of Semantic Correspondence},
  author    = {Cuttano, Claudia and Trivigno, Gabriele and Masone, Carlo and Roth, Stefan},
  booktitle = CVPR,
  year      = {2026}
}

@inproceedings{rombach2022high,
  title     = {High-Resolution Image Synthesis with Latent Diffusion Models},
  author    = {Rombach, Robin and Blattmann, Andreas and Lorenz, Dominik and Esser, Patrick and Ommer, Bj{\"o}rn},
  booktitle = CVPR,
  pages     = {10684--10695},
  year      = {2022}
}

@inproceedings{tumanyan2023plug,
  title     = {Plug-and-Play Diffusion Features for Text-Driven Image-to-Image Translation},
  author    = {Tumanyan, Narek and Geyer, Michal and Bagon, Shai and Dekel, Tali},
  booktitle = CVPR,
  pages     = {1921--1930},
  year      = {2023}
}

@inproceedings{zhang2023tale,
  title     = {A Tale of Two Features: Stable Diffusion Complements {DINO} for Zero-Shot Semantic Correspondence},
  author    = {Zhang, Junyi and Herrmann, Charles and Hur, Junhwa and Polan{\'i}a, Luisa F. and Jampani, Varun and Sun, Deqing and Yang, Ming-Hsuan},
  booktitle = NIPS,
  year      = {2023}
}

@inproceedings{mariotti2024spherical,
  title     = {Improving Semantic Correspondence with Viewpoint-Guided Spherical Maps},
  author    = {Mariotti, Octave and Mac Aodha, Oisin and Bilen, Hakan},
  booktitle = CVPR,
  year      = {2024}
}

@inproceedings{dunkel2025diy,
  title     = {Do It Yourself: Learning Semantic Correspondence from Pseudo-Labels},
  author    = {D{\"u}nkel, Olaf and Wimmer, Thomas and Theobalt, Christian and Rupprecht, Christian and Kortylewski, Adam},
  booktitle = ICCV,
  year      = {2025}
}

@article{liu2025partfield,
  title   = {{PartField}: Learning {3D} Feature Fields for Part Segmentation and Beyond},
  author  = {Liu, Minghua and Uy, Mikaela Angelina and Xiang, Donglai and Su, Hao and Fidler, Sanja and Sharp, Nicholas and Gao, Jun},
  journal = {arXiv preprint arXiv:2504.11451},
  year    = {2025}
}

@inproceedings{luo2023diffusion,
  title = {Diffusion hyperfeatures: Searching through time and space for semantic correspondence},
  booktitle = NIPS,
  author = {Luo, Grace and Dunlap, Lisa and Park, Dong Huk and Holynski, Aleksander and Darrell, Trevor},
  year = {2023}
}

@inproceedings{taniai2016joint,
  title     = {Joint Recovery of Dense Correspondence and Cosegmentation in Two Images},
  author    = {Taniai, Tatsunori and Sinha, Sudipta N. and Sato, Yoichi},
  booktitle = CVPR,
  pages     = {4246--4255},
  year      = {2016}
}

@inproceedings{Yi16,
  title={A scalable active framework for region annotation in 3D shape collections},
  author={Yi, Li and Kim, Vladimir G and Ceylan, Duygu and Shen, Wei and Yan, Mengyuan and Su, Hao and Lu, Cewu and Huang, Qixing and Sheffer, Alla and Guibas, Leonidas},
  booktitle=SIGGRAPHAsia,
  year={2016}
}

@misc{Min19SPair,
      title={SPair-71k: A Large-scale Benchmark for Semantic Correspondence}, 
      author={Min, Juhong and Lee, Jongmin and Ponce, Jean and Cho, Minsu},
      year={2019},
      eprint={1908.10543},
      archivePrefix={arXiv},
      primaryClass={cs.CV},
      url={https://arxiv.org/abs/1908.10543}, 
}

@article{Neverova20,
  title     = {Continuous Surface Embeddings for Deformable Shape Correspondence},
  author = {Neverova, Natalia and Novotny, David and Khalidov, Vasil and Szafraniec, Marc and Labatut, Patrick and Vedaldi, Andrea},  
  journal   = NIPS,
  year      = {2020}
}

@inproceedings{SHIC,
      title={SHIC: Shape-Image Correspondences with no Keypoint Supervision}, 
      author={Shtedritski, Aleksandar and Rupprecht, Christian and Vedaldi, Andrea},
      year={2024},
      booktitle=ECCV,
}

@InProceedings{Common3D,
    author    = {Sommer, Leonhard and D\"unkel, Olaf and Theobalt, Christian and Kortylewski, Adam},
    title     = {Common3D: Self-Supervised Learning of 3D Morphable Models for Common Objects in Neural Feature Space},
    booktitle = CVPR,
    month     = {June},
    year      = {2025},
    pages     = {6468-6479}
}

@article{Lowe04SIFT,
  author    = {David G. Lowe},
  title     = {Distinctive Image Features from Scale-Invariant Keypoints},
  journal   = IJCV,
  volume    = {60},
  number    = {2},
  pages     = {91--110},
  year      = {2004}
}

@article{Liu11SIFTFlow,
  author = {Liu, Ce and Yuen, Jenny and Torralba, Antonio},  
title     = {SIFT Flow: Dense Correspondence across Scenes and Its Applications},
  journal   = PAMI,
  volume    = {33},
  number    = {5},
  pages     = {978--994},
  year      = {2011}
}

@article{oquab2023dinov2,
  title={{DINOv2}: Learning Robust Visual Features without Supervision},
  author={Oquab, Maxime and Darcet, Timoth{\'e}e and Moutakanni, Th{\'e}o and Vo, Huy and Szafraniec, Marc and Khalidov, Vasil and Fernandez, Pierre and Haziza, Daniel and Massa, Francisco and El-Nouby, Alaaeldin and others},
  journal={arXiv preprint arXiv:2304.07193},
  year={2023}
}

@article{zhu2024densematcher,
      title={DenseMatcher: Learning 3D Semantic Correspondence for Category-Level Manipulation from a Single Demo},
      author={Zhu, Junzhe and Ju, Yuanchen and Zhang, Junyi and Wang, Muhan and Yuan, Zhecheng and Hu, Kaizhe and Xu, Huazhe},
      journal=ICLR,
      year={2025}
    }

@INPROCEEDINGS{canSurfMap2019abhinav,
  author={Kulkarni, Nilesh and Tulsiani, Shubham and Gupta, Abhinav},
  booktitle=ICCV, 
  title={Canonical Surface Mapping via Geometric Cycle Consistency}, 
  year={2019},
  volume={},
  number={},
  pages={2202-2211},
  doi={10.1109/ICCV.2019.00229}}

@INPROCEEDINGS{semalign3d2025,
  author={Wandel, Krispin and Wang, Hesheng},
  booktitle=CVPR, 
  title={SemAlign3D: Semantic Correspondence between RGB-Images through Aligning 3D Object-Class Representations}, 
  year={2025},
  volume={},
  number={},
  pages={1138-1147},
  doi={10.1109/CVPR52734.2025.00114}}

@misc{sam3d,
      title={{SAM 3D}: {3Dfy} Anything in Images},
      author={{SAM 3D Team} and Xingyu Chen and Fu-Jen Chu and Pierre Gleize and Kevin J Liang and Alexander Sax and Hao Tang and Weiyao Wang and Michelle Guo and Thibaut Hardin and Xiang Li and Aohan Lin and Jiawei Liu and Ziqi Ma and Anushka Sagar and Bowen Song and Xiaodong Wang and Jianing Yang and Bowen Zhang and Piotr Dollár and Georgia Gkioxari and Matt Feiszli and Jitendra Malik},
      year={2025},
      eprint={2511.16624},
      archivePrefix={arXiv},
      primaryClass={cs.CV},
      url={https://arxiv.org/abs/2511.16624}, 
}

@misc{sam3,
      title={{SAM 3}: Segment Anything with Concepts},
      author={Nicolas Carion and Laura Gustafson and Yuan-Ting Hu and Shoubhik Debnath and Ronghang Hu and Didac Suris and Chaitanya Ryali and Kalyan Vasudev Alwala and Haitham Khedr and Andrew Huang and Jie Lei and Tengyu Ma and Baishan Guo and Arpit Kalla and Markus Marks and Joseph Greer and Meng Wang and Peize Sun and Roman Rädle and Triantafyllos Afouras and Effrosyni Mavroudi and Katherine Xu and Tsung-Han Wu and Yu Zhou and Liliane Momeni and Rishi Hazra and Shuangrui Ding and Sagar Vaze and Francois Porcher and Feng Li and Siyuan Li and Aishwarya Kamath and Ho Kei Cheng and Piotr Dollár and Nikhila Ravi and Kate Saenko and Pengchuan Zhang and Christoph Feichtenhofer},
      year={2025},
      eprint={2511.16719},
      archivePrefix={arXiv},
      primaryClass={cs.CV},
      url={https://arxiv.org/abs/2511.16719},
}

@inproceedings{novum,
	 author  = {Artur Jesslen and Guofeng Zhang and Angtian Wang and Wufei Ma and Alan Yuille and Adam Kortylewski},
	 title   = {NOVUM: Neural Object Volumes for Robust Object Classification},
	 booktitle = ECCV,
	 year    = {2024}
 }

@inproceedings{OriAny2,
title={Orient Anything V2: Unifying Orientation and Rotation Understanding},
author={Wang, Zehan and Zhang, Ziang and Xu, Jiayang and Wang, Jialei and Pang, Tianyu and Du, Chao and Zhao, Hengshuang and Zhao, Zhou},
booktitle=NIPS,
year={2025},
}

@inproceedings{Zhang:2024:Telling,
  title={Telling Left from Right: {I}dentifying Geometry-Aware Semantic Correspondence},
  author={Zhang, Junyi and Herrmann, Charles and Hur, Junhwa and Chen, Eric and Jampani, Varun and Sun, Deqing and Yang, Ming-Hsuan},
  booktitle=CVPR,
  pages={3076--3085},
  year = {2024},
}

@inproceedings{Li:2024:Sd4match,
  title={{SD4Match}: {L}earning to Prompt {S}table {D}iffusion Model for Semantic Matching},
  author={Li, Xinghui and Lu, Jingyi and Han, Kai and Prisacariu, Victor Adrian},
  booktitle=CVPR,
  pages={27558--27568},
  year = {2024},
}

@inproceedings{Mariotti:2025:Jamais,
  title={Jamais {V}u: {E}xposing the Generalization Gap in Supervised Semantic Correspondence},
  author={Mariotti, Octave and Du, Zhipeng and Bhalgat, Yash and Mac Aodha, Oisin and Bilen, Hakan},
  booktitle=NIPS,
  volume={38},
  year=2025
}

@inproceedings{donati2020deep,
  title={Deep geometric functional maps: Robust feature learning for shape correspondence},
  author={Donati, Nicolas and Sharma, Abhishek and Ovsjanikov, Maks},
  booktitle=CVPR,
  pages={8592--8601},
  year={2020}
}

@inproceedings{wang2024gs,
  title={Gs-pose: Category-level object pose estimation via geometric and semantic correspondence},
  author={Wang, Pengyuan and Ikeda, Takuya and Lee, Robert and Nishiwaki, Koichi},
  booktitle=ECCV,
  pages={108--126},
  year={2024},
  organization={Springer}
}

@inproceedings{dutt2024diffusion,
  title={Diffusion 3d features (diff3f): Decorating untextured shapes with distilled semantic features},
  author={Dutt, Niladri Shekhar and Muralikrishnan, Sanjeev and Mitra, Niloy J},
  booktitle=CVPR,
  pages={4494--4504},
  year={2024}
}

@article{ovsjanikov2012functional,
  title={Functional maps: a flexible representation of maps between shapes},
  author={Ovsjanikov, Maks and Ben-Chen, Mirela and Solomon, Justin and Butscher, Adrian and Guibas, Leonidas},
  journal=TOG,
  volume={31},
  number={4},
  pages={1--11},
  year={2012},
  publisher={ACM New York, NY, USA}
}

@article{ham2017proposal,
  title   = {Proposal Flow: Semantic Correspondences from Object Proposals},
  author  = {Ham, Bumsub and Cho, Minsu and Schmid, Cordelia and Ponce, Jean},
  journal = PAMI,
  volume  = {40},
  number  = {7},
  pages   = {1711--1725},
  year    = {2017}
}

@article{aberman2018neural,
  title = {Neural best-buddies: Sparse cross-domain correspondence},
  journal = TOG,
  author = {Aberman, Kfir and Liao, Jing and Shi, Mingyi and Lischinski, Dani and Chen, Baoquan and Cohen-Or, Daniel},
  year = {2018}
}

@inproceedings{zhou2016learning,
  title = {Learning Dense Correspondence via 3D-Guided Cycle Consistency},
  author = {Zhou, Tinghui and Krahenbuhl, Philipp and Aubry, Mathieu and Huang, Qixing and Efros, Alexei A.},
  booktitle = CVPR,
  year = {2016}
}

@inproceedings{kim2022semi,
  title = {Semi-Supervised Learning of Semantic Correspondence with Pseudo-Labels},
  author = {Kim, Jiwon and Ryoo, Kwangrok and Seo, Junyoung and Lee, Gyuseong and Kim, Daehwan and Cho, Hansang and Kim, Seungryong},
  booktitle = CVPR,
  year = {2022}
}

@inproceedings{li2021probabilistic,
  title = {Probabilistic Model Distillation for Semantic Correspondence},
  author = {Li, Xin and Fan, Deng-Ping and Yang, Fan and Luo, Ao and Cheng, Hong and Liu, Zicheng},
  booktitle = CVPR,
  year = {2021}
}

@inproceedings{huang2023weakly,
  title = {Weakly Supervised Learning of Semantic Correspondence through Cascaded Online Correspondence Refinement},
  author = {Huang, Yiwen and Sun, Yixuan and Lai, Chenghang and Xu, Qing and Wang, Xiaomei and Shen, Xuli and Ge, Weifeng},
  booktitle = ICCV,
  year = {2023}
}

@inproceedings{hedlin2023unsupervised,
  title = {Unsupervised Semantic Correspondence Using Stable Diffusion},
  author = {Hedlin, Eric and Sharma, Gopal and Mahajan, Shweta and Isack, Hossam and Kar, Abhishek and Tagliasacchi, Andrea and Yi, Kwang Moo},
  booktitle = NIPS,
  year = {2023}
}

@inproceedings{tang2023emergent,
  title = {Emergent Correspondence from Image Diffusion},
  author = {Tang, Luming and Jia, Menglin and Wang, Qianqian and Phoo, Cheng Perng and Hariharan, Bharath},
  booktitle = NIPS,
  year = {2023}
}

@inproceedings{fundel2024distillation,
  title = {Distillation of Diffusion Features for Semantic Correspondence},
  author = {Fundel, Frank and Schusterbauer, Johannes and Hu, Vincent Tao and Ommer, Bjorn},
  booktitle = WACV,
  year = {2025}
}

@article{xue2025matcha,
  title = {{MATCHA}: Towards Matching Anything},
  author = {Xue, Fei and Elflein, Sven and Leal-Taixe, Laura and Zhou, Qunjie},
  journal = {arXiv preprint arXiv:2501.14945},
  year = {2025}
}

@inproceedings{Yu:2021:AP10k,
  title = {{AP}-10K: A Benchmark for Animal Pose Estimation in the Wild},
  author = {Yu, Hang and Xu, Yufei and Zhang, Jing and Zhao, Wei and Guan, Ziyu and Tao, Dacheng},
  booktitle = NIPS,
  year = {2021}
}

@inproceedings{hartwig2025geco,
  title={GECO: Geometrically Consistent Embedding with Lightspeed Inference},
  author={Hartwig, Regine and Muhle, Dominik and Marin, Riccardo and Cremers, Daniel},
  booktitle=ICCV,
  pages={9309--9319},
  year={2025}
}

@inproceedings{chic3po,
  title={C3PO: Canonicalization of 3D Pose from Partial Views with Generalizable Correspondence Features},
  author={Chi, Yu and Sommer, Leonhard and D{\"u}nkel, Olaf and Muhle, Dominik and Cremers, Daniel and Theobalt, Christian and Kortylewski, Adam},
  booktitle=ThreeDV,
  year={2026}
}
}
\appendix
\maketitlesupplementary
\renewcommand\thesection{\Alph{section}}
\numberwithin{equation}{section}
\numberwithin{figure}{section}
\numberwithin{table}{section}
\renewcommand{\thefigure}{\thesection\arabic{figure}}
\renewcommand{\thetable}{\thesection\arabic{table}}

\newcommand{\additem}[2]{%
\item[\textbf{(\ref{#1})}] 
    \textbf{#2} \dotfill\makebox{\textbf{\pageref{#1}}
    }
}

\newcommand{\myindent}{.5em}
\newcommand{\addsubitem}[2]{%
\vspace{.5em}
    \textbf{(\ref{#1})}
        \hspace{\myindent} #2 \\    
}

\newcommand{\adddescription}[1]{\vspace{.1em}
\begin{adjustwidth}{0cm}{0cm}
#1
\end{adjustwidth}
}
\setlist[itemize]{noitemsep,leftmargin=*,topsep=0em}
\setlist[enumerate]{noitemsep,leftmargin=*,topsep=0em}

\noindent This supplement is organized as follows. \cref{supp:pseudo_gt_pose} provides implementation details for the 3D reconstruction and pose canonicalization pipeline. \cref{supp:pseudo_gt_correspondences} gives additional details on feature fusion, pseudo-label generation, and geodesic filtering. \cref{supp:additional_results_viz} reports per-category results and additional qualitative visualizations. \cref{supp:llm} discusses reproducibility and the use of LLM assistance in writing this paper.

\vspace{0.3in}

\begin{enumerate}[label={({\arabic*})}, topsep=1em, itemsep=2em]
    \additem{supp:pseudo_gt_pose}{Pseudo-groundtruth via foundation models}
    \additem{supp:pseudo_gt_correspondences}{Correspondence pseudo-annotations}
    \additem{supp:additional_results_viz}{Additional results and visualizations}
    \additem{supp:llm}{Reproducibility and LLM assistance}
\end{enumerate}

\setlength{\parskip}{.5em}
\clearpage
\section{Pseudo-groundtruth via foundation models}\label{supp:pseudo_gt_pose}

This section provides additional details on the 3D reconstruction and canonicalization pipeline summarized in \cref{sec:meth:pseudo_gt_pose}. The full pipeline is illustrated in \cref{supp:fig:pipeline-data}. Starting from a single image, we (i) extract a 2D instance mask and reconstruct an object-centric mesh using foundation models, (ii) refine the mesh pose via a two-phase render-and-compare optimization, and (iii) resolve any residual yaw ambiguity by comparing rendered views against estimated orientations.

\paragraph{2D Mask and 3D Mesh Initialization}
To extract instance masks with SAM3~\citep{sam3}, we prompt the model with the category label provided by SPair-71k. These prompts improve mask quality and reduce failure cases, but are not strictly required: masks can be obtained without them, at the cost of additional noise in the downstream pipeline. We validate this choice empirically: a simple baseline using DINOv2 CLS token embeddings with kNN classification achieves $\sim 99\%$ accuracy on the category classification task. Failure cases occur primarily when multiple objects occupy a single image (affecting <1\% of instances), representing a negligible impact on overall performance.
We consider this a reasonable choice since neither the bounding box nor the category label constitutes additional human annotation beyond what the dataset already provides, and both can be obtained automatically with off-the-shelf object detectors if needed.

\paragraph{Render-and-compare pose refinement}
The interior-coverage reward ($\lambda > 1$ in \cref{eq:dt_loss}) is critical in cases of strong occlusion. Without it, we found that the optimization sometimes escapes the distance-transform penalty by pushing the rendered silhouette entirely outside the image — avoiding the loss rather than solving it (\eg, the partially occluded chair in \cref{supp:fig:pipeline-data}). The interior-coverage term counteracts this by rewarding rendered mass that lands inside the observed mask, preventing the degenerate solution. We additionally apply a strong penalty whenever more than 25\% of the rendered silhouette falls outside the image boundary.

We set $\lambda=4$ and dilate the observed mask by $r=4$ pixels before computing the distance-transform fields (see \cref{eq:dist_transform}), providing tolerance for coarse mesh boundaries. We optimize log-scale and translation jointly with Adam, using separate learning rates $\text{lr}_\text{scale}=0.05$ and $\text{lr}_\text{trans}=0.02$. The higher learning rate for scale biases the optimization toward correcting the dominant error (scale mismatch) rather than compensating with depth drift, which is harmless for 2D reprojection quality but can destabilize the geometry. We run the distance-transform phase for 100 gradient steps, then switch to soft-IoU refinement for a further 50 steps to tighten the final alignment.

\paragraph{Yaw Canonicalization statistics}
Following the canonicalization verification procedure described in \cref{sec:meth:pseudo_gt_pose}, we applied discrete orientation corrections to a small subset of the refined meshes. Excluding meshes marked as wrong, 79 out of 1,319 instances required a non-zero rotation, corresponding to 5.99\% of the dataset. These corrections indicate cases where the estimated orientation differed from the target canonical pose by one of the allowed discrete rotations.

All corrections were rotations around the \(y\)-axis: 34 meshes required a \(270^\circ\) rotation, 24 required a \(90^\circ\) rotation, and 15 required a \(180^\circ\) rotation. The corrections were distributed across both splits, with 59 rotated meshes in the training set and 20 in the validation set. The most frequently affected classes were bus, boat, train, and cow, with 23, 15, 8, and 7 corrected meshes, respectively.

\clearpage
\section{Correspondence pseudo-annotations}\label{supp:pseudo_gt_correspondences}

This section provides additional details on the correspondence pseudo-annotation pipeline described in \cref{sec:meth:pseudo_gt_corresp}. We first give further analysis of the PartField features used in our feature fusion, including PCA visualizations and rasterization details (\cref{supp:feature-partfield}). We then detail the feature fusion weight search and justify the square-root weighting scheme (\cref{supp:feature-weights-search}).

\subsection{PartField Features}\label{supp:feature-partfield}

\begin{figure}[h]
    \centering
    \begin{subfigure}[t]{0.45\linewidth}
        \centering
        \includegraphics[width=\linewidth]{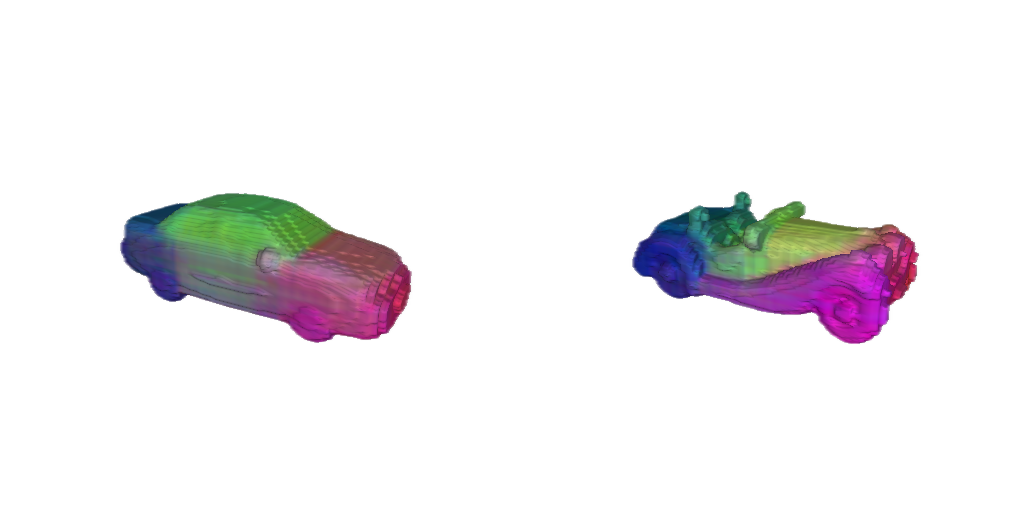}
        \caption{PCA projection across two car instances.}
        \label{supp:fig:partfield_pca_car}
    \end{subfigure}
    \hfill
    \begin{subfigure}[t]{0.45\linewidth}
        \centering
        \includegraphics[width=\linewidth]{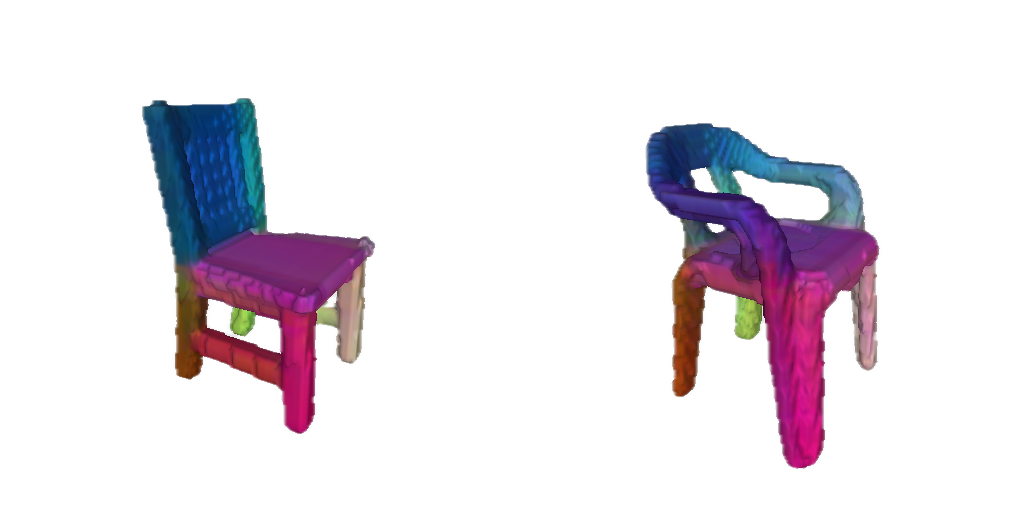}
        \caption{PCA projection across two chair instances.}
        \label{supp:fig:partfield_pca_chair}
    \end{subfigure}

    \caption{
    \textbf{PCA visualizations of PartField features.}
    We project PartField features to RGB using PCA and visualize them on pairs of object instances. Consistent colors within individual parts indicate that the features are spatially coherent, while similar colors across instances suggest that corresponding geometric parts, such as chair legs or car body regions, are mapped to nearby feature representations.
    }
    \label{supp:fig:partfield_pca}
\end{figure}

\begin{figure}[h]
    \centering
    \begin{subfigure}[t]{0.45\linewidth}
        \centering
        \includegraphics[width=\linewidth]{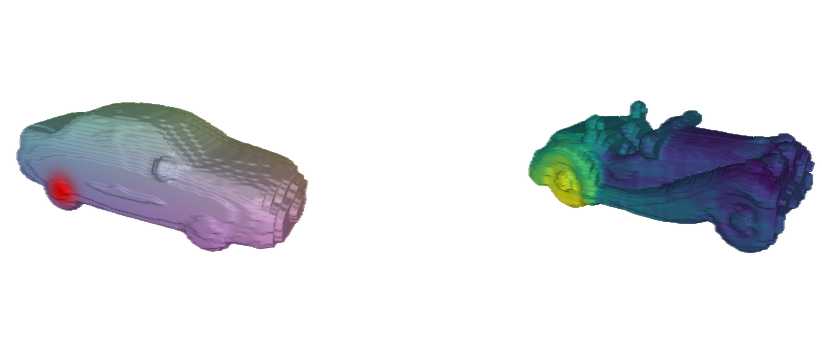}
        \caption{Highest similarity is correctly localized to the queried right-rear wheel.}
        \label{supp:fig:partfield_similarity_wheel}
    \end{subfigure}
    \hfill
    \begin{subfigure}[t]{0.45\linewidth}
        \centering
        \includegraphics[width=\linewidth]{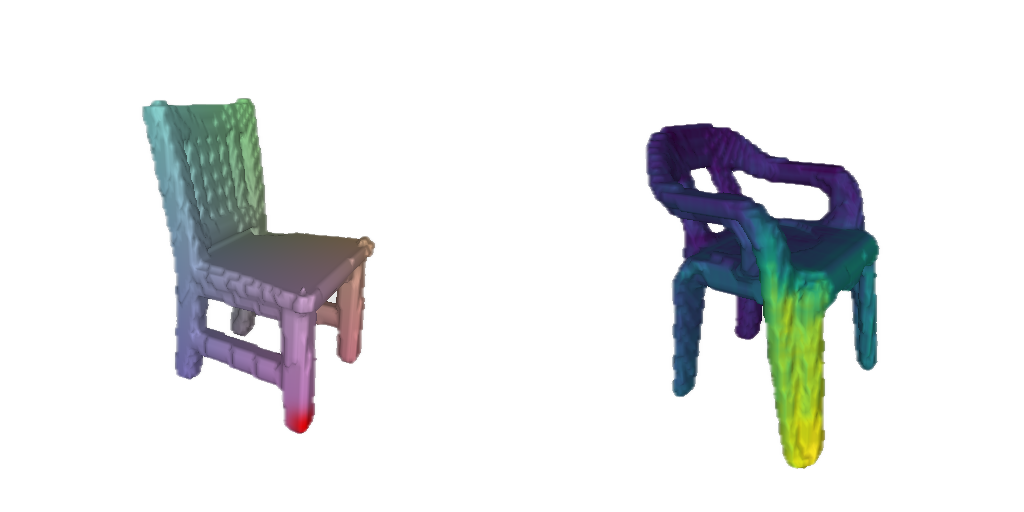}
        \caption{Highest similarity is correctly localized to the queried right-front chair leg.}
        \label{supp:fig:partfield_similarity_chair}
    \end{subfigure}

    \caption{
    \textbf{PartField features reduce repeated-part and symmetry ambiguities.}
    For each example, the left mesh shows the query point in red, and the right mesh shows the cosine-similarity heatmap induced by the queried PartField feature. In the car example, similarity concentrates on the queried wheel rather than activating all repeated wheels. In the chair example, the response remains localized to the corresponding leg, avoiding front/back and left/right confusion. This suggests that PartField similarities are anchored in geometric context rather than only in semantic part identity.
    }
    \label{supp:fig:partfield_similarity}
\end{figure}

We visualize PartField features~\cite{liu2025partfield} using PCA projections (\cref{supp:fig:partfield_pca}) and query-based similarity heatmaps (\cref{supp:fig:partfield_similarity}). These visualizations show that PartField features are spatially coherent within semantic parts while remaining discriminative across repeated or symmetric structures.

\paragraph{Rasterization details}
To rasterize PartField vertex features into the image plane, we first render the reconstructed 3D mesh into the image using the estimated pose, obtaining a mapping from each pixel to its corresponding 3D point on the mesh. We then assign to each pixel the PartField feature of its corresponding 3D point, effectively projecting the 3D-aware features into the 2D image space. This process allows us to leverage the geometric context captured by PartField features while maintaining alignment with the original image, enabling more accurate correspondence estimation.

\subsection{Feature fusion}\label{supp:feature-weights-search}
\begin{figure}[h]
    \centering
    \includegraphics[width=\linewidth]{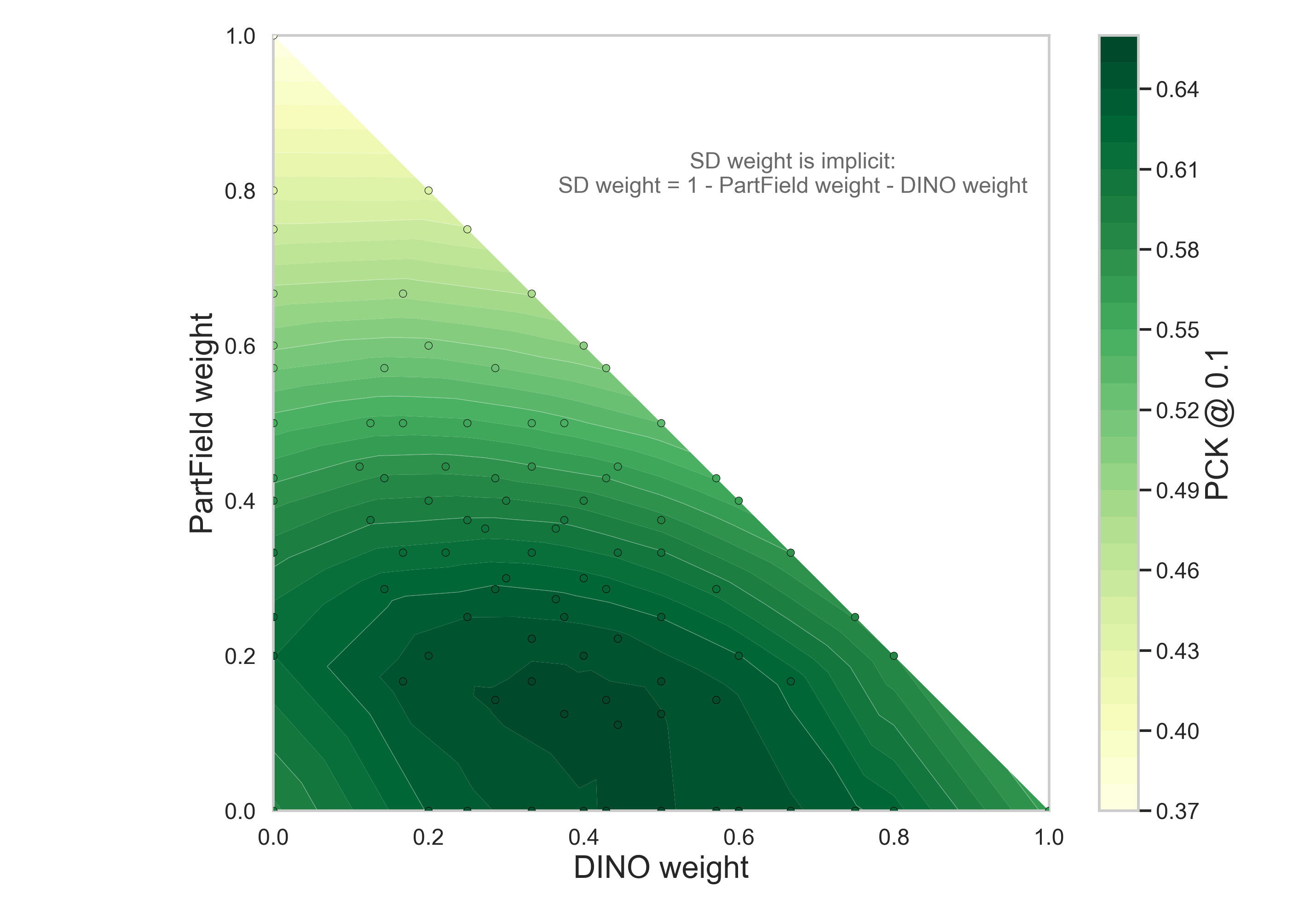}
    \caption{
    \textbf{Feature fusion weight search.}
    PCK@0.10 of pseudo-correspondences before filtering on the SPair-71k validation set, as a function of the SD weight $\alpha$ and DINOv2 weight $\beta$ (the PartField weight $\gamma = 1 - \alpha - \beta$ is determined by the other two).
    Multiple combinations achieve similar peak performance; we use $\alpha=1/2$, $\beta=1/3$, $\gamma=1/6$ as our default.
    }
    \label{supp:fig:weight_search}
\end{figure}

We select the fusion weights $\alpha$, $\beta$, and $\gamma$ from \cref{eq:fused_features} via a grid search on the SPair-71k validation set, sweeping $\alpha$ and $\beta$ in increments of $1/6$ with $\gamma = 1 - \alpha - \beta$, and measuring \pckTen{} of unfiltered predicted correspondences.
As shown in \cref{supp:fig:weight_search}, several weight combinations reach similar peak performance, confirming that the method is not too sensitive to the exact weighting and all features provide some contributions.
Among these, we select $\alpha=1/2$, $\beta=1/3$, and $\gamma=1/6$: configurations that upweight PartField ($\gamma$) tend to yield better downstream performance after adapter training, consistent with PartField resolving the geometrically ambiguous correspondences that provide the most informative supervision signal.

\paragraph{Square-root fusion weights}
Let $\widehat{\Feature}_{\text{SD}}$, $\widehat{\Feature}_{\text{DINO}}$, and
$\widehat{\Feature}_{\text{PF}}$ denote the independently L2-normalized feature
vectors of any two candidate points. The dot product between their fused features is
\begin{align}
\Feature_{\text{fused}}^\top\Feature_{\text{fused}}'
&=\sqrt{\alpha}\,\widehat{\Feature}_{\text{SD}}^\top
\sqrt{\alpha}\,\widehat{\Feature}_{\text{SD}}'
+
\sqrt{\beta}\,\widehat{\Feature}_{\text{DINO}}^\top
\sqrt{\beta}\,\widehat{\Feature}_{\text{DINO}}'
+
\sqrt{\gamma}\,\widehat{\Feature}_{\text{PF}}^\top
\sqrt{\gamma}\,\widehat{\Feature}_{\text{PF}}'\\
&=
\alpha\,
\widehat{\Feature}_{\text{SD}}^\top
\widehat{\Feature}_{\text{SD}}'
+
\beta\,
\widehat{\Feature}_{\text{DINO}}^\top
\widehat{\Feature}_{\text{DINO}}'
+
\gamma\,
\widehat{\Feature}_{\text{PF}}^\top
\widehat{\Feature}_{\text{PF}}' .
\end{align}
Since each feature source is L2-normalized independently, each dot product is
exactly the cosine similarity within that feature space. Therefore, the cosine
similarity in the concatenated fused space is equivalent to a weighted average
of the cosine similarities computed independently for each feature source. The
square roots appear because the weights are applied to both vectors before
taking the dot product, e.g. $\sqrt{\alpha}\sqrt{\alpha}=\alpha$.
\clearpage
\section{Additional results and visualizations}\label{supp:additional_results_viz}

\subsection{Additional implementation details}\label{supp:sec:implementation_details}
\paragraph{Compute}
Unless stated otherwise, all reported runtimes are measured on a single NVIDIA L40 GPU with 40\,GB of memory; our pipeline is also compatible with smaller memory budgets. The canonicalized 3D object reconstruction takes $12.42$\,s per object on average. Computing the pseudo-labels for the full SPair-71k training set ($\sim$53k pairs) takes roughly $18$\,h end-to-end, including SD, DINO, and PartField feature extraction, rasterization of PartField descriptors, cyclic consistency, and geodesic filtering for each image pair. Note that this pipeline could benefit from further optimization and parallelization to reduce runtime with minimal work.
Training the adapter for 200k iterations takes about $4$\,h on a single GPU.

\subsection{Additional results}\label{supp:sec:additional_results}
We provide additional results complementing those in the main paper. In particular, we extend the benchmark tables to include \textit{Supervised} methods in \cref{supp:tab:full_table}, which were omitted from the main paper for space, and report per-category PCK results on SPair-71k in \cref{supp:tab:Spair_all_cats}.
We exclude the \textit{weakly supervised without human annotations} variant of Telling Left from Right (Geo-SC)~\citep{Zhang:2024:Telling} from the main paper, as it reports PCK normalized per keypoint rather than per image, making direct comparison unreliable. For completeness, per-category results for Geo-SC are included in \cref{supp:tab:Spair_all_cats}.
\begin{table*}[t]
\centering
\caption{\textbf{Evaluation on standard benchmarks.} 
Per-image PCK (\%, $\uparrow$) at multiple thresholds on SPair-71k (test set and Geo-Aware subset), AP-10K and SpairU. 
$^\dag$ Results obtained from the official checkpoint.
`\text{--}' indicates missing numbers. Best per method type is shown in \textbf{bold}. }
\label{supp:tab:full_table}
\setlength{\tabcolsep}{3pt}
\footnotesize
\begin{tabularx}{\linewidth}{@{}lr|ccc|ccc|ccc|ccc@{}}
\toprule
 && \multicolumn{3}{c}{\textbf{SPair-71k}} 
 & \multicolumn{3}{c}{\textbf{SPair-Geo-Aware}} 
& \multicolumn{3}{c}{\textbf{AP-10K} (0.10)} 
& \multicolumn{3}{c@{}}{\textbf{SpairU}} \\
\cmidrule(lr){3-5}\cmidrule(lr){6-8}\cmidrule(lr){9-11}\cmidrule(lr){12-14}
\textbf{Method} &&
0.01 & 0.05 & 0.10 &
0.01 & 0.05 & 0.10 
& I.S. 
& C.S.  
& C.F.
& 0.01 & 0.05 & 0.10  \\
\midrule
\multicolumn{11}{@{}l}{\textit{Supervised}} \\
\myrowcolour
~~DHF&\citet{luo2023diffusion}
& 8.7 & 50.2 & 64.9
& 8.0 & 45.8 & 62.7
& 62.7 
& 60.0 
& 47.8 
& \text{--} & \text{--} & \text{--} \\
~~SD{+}DINOv2&\citet{zhang2023tale}
& 9.6 & 57.7 & 74.6 
& 9.9 & 57.0 & 77.0
& 77.0 
& 74.0 
& 65.8 
& \text{--} & \text{--} & \text{--} \\
\myrowcolour
~~GECO&\citet{hartwig2025geco}
& 14.2 & 59.6 & 73.6 
& \text{--} & \text{--} & \text{--}
& 82.5 
& 81.2 
& 76.6 
& \text{--} & \text{--} & 55.2 \\
~~Jamais Vu&\citet{Mariotti:2025:Jamais}
& 20.5 & 71.9 & 82.5 
& \text{--} & \text{--} & \text{--}
& \text{--} 
& \text{--} 
& \text{--} 
& \text{--} & \text{--} & 62.4 \\
\myrowcolour
~~Geo\text{-}SC&\citet{Zhang:2024:Telling}
& 21.7 & 72.8 & 83.2 
& \text{--} & \text{--} & \text{--}
& 87.7 
& 85.9 
& 78.5
& \text{--} & \text{--} & 56.9 \\
~~SemAlign3D&\citet{semalign3d2025}
& 15.8 & 77.5 & \textbf{88.9} 
& \text{--} & \text{--} & \text{--} 
& \text{--} & \text{--} & \text{--} 
& \text{--} & \text{--} & \text{--}\\
\myrowcolour
~~MARCO&\citet{cuttano2026marco}
& \textbf{27.0} & \textbf{77.6} & 87.2 
& \textbf{22.8}$^\dag$ & \textbf{76.8}$^\dag$ & \textbf{87.5}$^\dag$ 
& \textbf{89.1} 
& \textbf{88.3} 
& \textbf{83.4} 
& $5.0^\dag$ & $42.7^\dag$ & \textbf{67.5} \\

\midrule
\multicolumn{11}{@{}l}{\textit{Unsupervised}} \\
\myrowcolour
~~DINOv2+NN&\citet{zhang2023tale}
& 6.3 & 38.4 & \textbf{53.9} 
& \textbf{3.4} & \textbf{28.2} & 42.0
& \textbf{60.9}
& \textbf{57.3}
& \textbf{47.4} 
& \text{--} & \text{--} & \textbf{54.9} \\
~~DIFT&\citet{tang2023emergent}
& \textbf{7.2} & \textbf{39.7} & 52.9 
& \textbf{3.4} & \textbf{28.2} & \textbf{42.5}
& 50.3 
& 46.0 
& 35.0 
& \text{--} & \text{--} & 47.4 \\
\midrule
\multicolumn{11}{@{}l}{\textit{Weakly Supervised with human annotations}} \\
\myrowcolour
~~Spherical Map.& \citet{mariotti2024spherical}
& \text{8.4} & \text{48.2} & \text{64.4}
& \text{--} & \text{--} & \text{--}
& 65.4 & 63.1 & 51.0
& \text{--} & \text{--} & 61.0\\
~~DIY-SC& \citet{dunkel2025diy}
& \textbf{10.1} & \textbf{53.8} & \textbf{71.6} 
& 7.7 & 47.7 & 67.5
& \textbf{70.6}
& \textbf{69.8}
& \textbf{57.8} 
& 5.4 & 44.0 & \textbf{67.9} \\

\midrule
\multicolumn{11}{@{}l}{\textit{Weakly Supervised without human annotations}} \\

\myrowcolour
~~SD{+}DINOv2&\citet{zhang2023tale}
& 7.9 & 44.7 & 59.9 
& 5.3 & 34.5 & 49.3
& 62.9 
& 59.3 
& 48.3 
& \text{--} & \text{--} & 59.4\\
~~DIY-SC+OriAny&\citet{dunkel2025diy} 
& 9.5 & 51.2 & 69.6
& 6.9 & 45.7 & 65.8
& 69.3
& 66.8
& 54.0
& 5.2 & 43.1 & 66.3 \\
\myrowcolour
~~\textbf{\ours (Ours)}& 
& \textbf{10.2} & \textbf{54.8} & \textbf{73.0}
& \textbf{7.8} & \textbf{50.1} & \textbf{70.8}
& \textbf{69.6}
& \textbf{68.5}
& \textbf{56.9}
& \textbf{5.6} & \textbf{43.5} & \textbf{67.3} \\
\bottomrule
\end{tabularx}
\end{table*}

Per-category results are reported in \cref{supp:tab:Spair_all_cats} (\textbf{per-keypoint \pckTen}). The pattern of gains is consistent with our central hypothesis: the largest improvements over DIY-SC+OriAny (and even DIY-SC which was trained with human supervision) occur in rigid, man-made categories with strong geometric symmetry — bus ($+10.8$), tv/monitor ($+9.8$), bottle ($+8.8$), car ($+6.9$), train ($+6.2$), motorcycle ($+5.1$), and chair ($+4.0$). These are precisely the categories where 2D features tend to confuse symmetric sides or visually similar parts, and where PartField descriptors provide the strongest disambiguating signal. By contrast, non-rigid animal categories such as sheep ($-2.7$), cat ($-1.5$), and cow ($-1.7$) show slight regressions, which is expected: PartField is trained with a part-level contrastive objective on rigid objects and generalizes less reliably to deformable shapes. Potted plant similarly shows a marginal decrease ($-0.6$), likely because SAM3D reconstructs the pot and plant as a single merged shape, whereas evaluation keypoints typically land on the pot alone.

\begin{table}[t]
    \caption{\textbf{Per-category PCK@0.1 scores (\textit{per-keypoint}) on SPair-71k}. Gains are largest for rigid, man-made categories with strong geometric symmetry (bus, tv/monitor, car, motorcycle), where PartField features resolve left--right and repeated-part ambiguities. Non-rigid categories such as animals show little to no improvement.
     }
    \centering
    \resizebox{\textwidth}{!}{
    \begin{tabular}{l|rrrrrrrrrrrrrrrrrr|r}
             & \faIcon{plane} & \faIcon{bicycle} & \faIcon{crow} & \faIcon{ship} & \faIcon{wine-bottle} & \faIcon{bus} & \faIcon{car} & \faIcon{cat} & \faIcon{chair} & \Cow  & \faIcon{dog} & \faIcon{horse} & \faIcon{motorcycle} & \faIcon{walking} & \Plant & \Sheep & \faIcon{train} & \faIcon{tv} & avg\\ 
        \midrule
        \multicolumn{20}{@{}l}{\textit{Supervised}} \\
        \myrowcolour
        ~~MARCO
            & 93.7 & 79.8 & 96.9 & 74.7 & 75.4 & 95.2 & 91.9 & 94.8 & 87.5 & 96.5 & 91.2 & 90.3 & 87.6 & 63.x & 29.x & 63.x & 51.x & 29.x & 57.x \\

        \midrule
        \multicolumn{20}{@{}l}{\textit{Unsupervised}} \\
        \myrowcolour
        ~~DINOv2+NN
            & 72.7 & 62.0 & 85.2 & 41.3 & 40.4 & 52.3 & 51.5 & 71.1 & 36.2 & 67.1 & 64.6 & 67.6 & 61.0 & 68.2 & 30.7 & 62.0 & 54.3 & 24.2 & 55.6 \\
        
        ~~DIFT 
            & 63.5 & 54.5 & 80.8 & 34.5 & 46.2 & 52.7 & 48.3 & 77.7 & 39.0 & 76.0 & 54.9 & 61.3 & 53.3 & 46.0 & 57.8 & 57.1 & 71.1 & 63.4 & 57.7 \\

        \midrule
        \multicolumn{20}{@{}l}{\textit{Weakly Supervised with human annotations}} \\
        \myrowcolour
        ~~Spherical Mapper  
            &75.3 & 63.8 & 87.7 & 48.2 & 50.9 & 74.9 & 71.1 & 81.7 & 47.3 & 81.6 & 66.9 & 73.1 & 65.4 & 61.8 & 55.5 & 70.2 & 75.0 & 58.5 & 67.8 \\
        ~~DIY-SC
            & \textbf{77.2} & \textbf{69.1} & \textbf{90.8} & \textbf{54.2} & \textbf{57.9} & \textbf{83.7} & \textbf{77.5} & \textbf{86.5} & \textbf{53.1} & \textbf{86.7} & \textbf{73.1} & \textbf{78.5} & \textbf{72.5} & \textbf{74.0} & \textbf{73.5} & \textbf{76.0} & \textbf{77.2} & \textbf{69.5} & \textbf{74.4} \\
        \midrule
        \multicolumn{20}{@{}l}{\textit{Weakly Supervised without human annotations}} \\
        \myrowcolour
        ~~SD+DINOv2  
            & 73.0 & 64.1 & 86.4 & 40.7 & 52.9 & 55.0 & 53.8 & 78.6 & 45.5 & 77.3 & 64.7 & 69.7 & 63.3 & 69.2 & 58.4 & 67.6 & 66.2 & 53.5 & 64.0 \\
        ~~Geo\text{-}SC
            & \textbf{78.0} & 66.4 & 90.2 & 44.5 & 60.1 & 66.6 & 60.8 & 82.7 & 53.2 & 82.3 & 69.5 & 75.1 & 66.1 & 71.7 & 58.9 & 71.6 & 83.8 & 55.5 & 69.6 \\
        ~~DIY-SC+OriAny
            & 76.1 & 65.9 & 90.4 & 52.2 & 57.3 & 75.7 & 75.3 & \textbf{85.0} & 52.8 & \textbf{86.3} & 71.4 & \textbf{78.3} & 69.9 & \textbf{73.5} & \textbf{69.2} & \textbf{75.0} & 76.7 & 69.6 & 72.9 \\
        ~~\ours (Ours) 
            & 77.6 & \textbf{70.3} & \textbf{90.4} & \textbf{54.8} & \textbf{66.1} & \textbf{86.5} & \textbf{82.2} & 83.5 & \textbf{56.8} & 84.6 & \textbf{72.6} & 77.8 & \textbf{75.0} & 72.5 & 68.6 & 72.3 & \textbf{82.9} & \textbf{79.4} & \textbf{76.3}\\

        \bottomrule
    \end{tabular}
    }
    \label{supp:tab:Spair_all_cats}
\end{table}

\subsection{Comparison of pseudo-annotations}\label{supp:comparison_filtered}
\begin{figure}[ht]
    \centering
    \begin{minipage}[t]{0.49\linewidth}
        \centering
        \includegraphics[width=\linewidth]{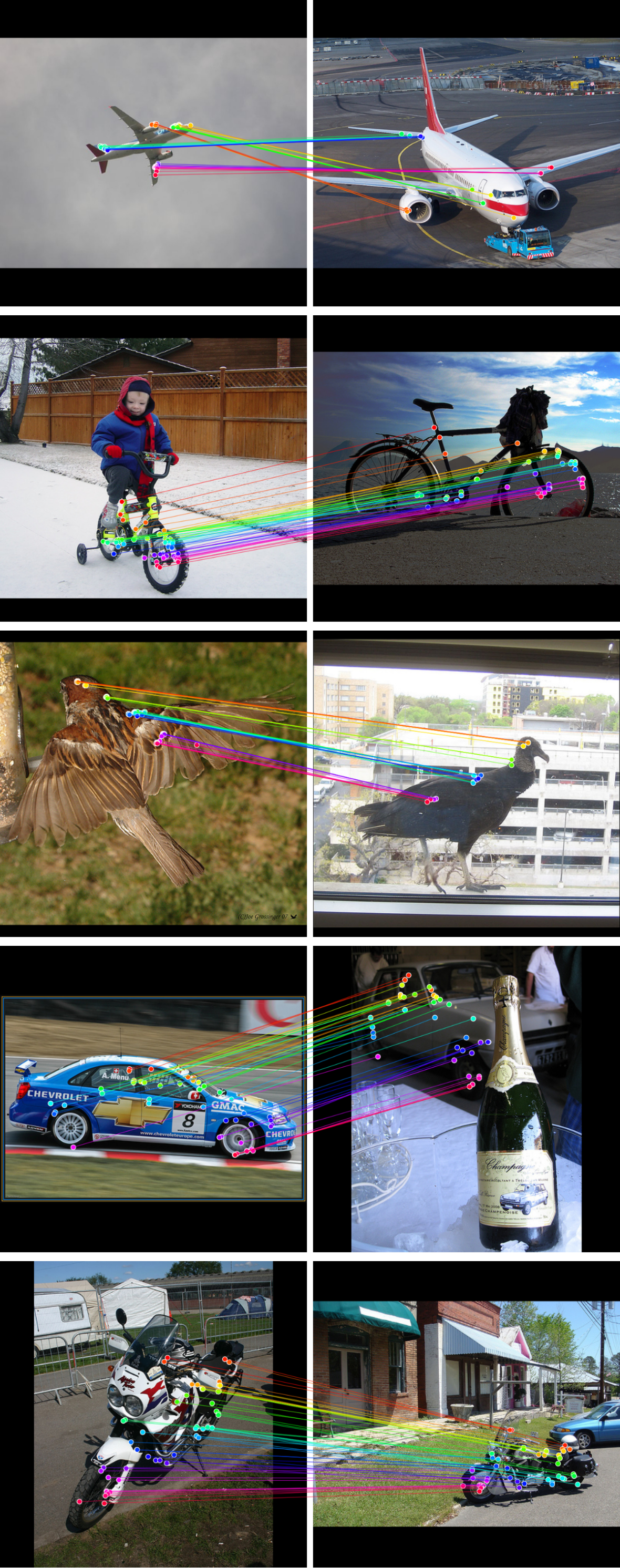}
        \subcaption{\centering\textbf{\ours.}}
    \end{minipage}
    \hfill
    \begin{minipage}[t]{0.49\linewidth}
        \centering
        \includegraphics[width=\linewidth]{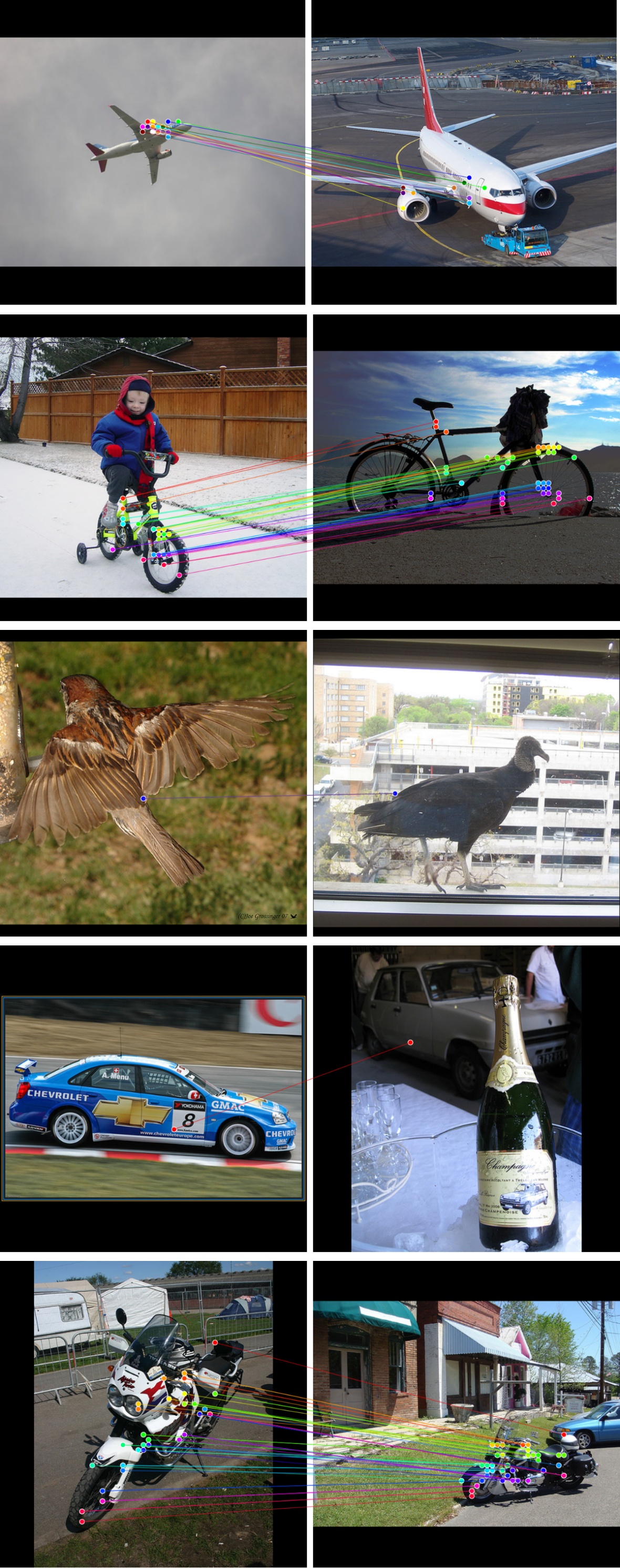}
        \subcaption{\centering\textbf{DIY-SC.}}
    \end{minipage}
    \caption{\textbf{Qualitative pseudo-annotations.} We visualize pseudo-ground-truth annotations from \ours{} and DIY-SC. \ours{} produces denser and more geometrically consistent pseudo-annotations.}
    \label{supp:fig:pseudo-annot-quali}
\end{figure}

\cref{supp:fig:pseudo-annot-quali} extends the qualitative comparison from the main paper with additional examples. Across object categories, \ours{} consistently produces denser pseudo-annotations that cover a larger fraction of the object surface, while remaining geometrically consistent and free from left-right ambiguities. By contrast, DIY-SC pseudo-labels are sparser and more prone to symmetric confusions, as its spherical geometric prior cannot resolve instance-level structure. These qualitative differences are directly reflected in the quantitative gains on SPair-Geo-Aware, which specifically targets symmetric and repeated-part correspondences.
\clearpage
\section{Reproducibility and LLM assistance}\label{supp:llm}
To ensure full reproducibility of our work, we will release all code and data used in this paper. The complete processing pipeline, including scripts for dataset preparation will be made publicly available on \href{https://github.com/GenIntel/3D-SC}{\faGithub\texttt{/GenIntel/3D-SC}}. Our training and inference code for the proposed model is provided in the same repository, together with configuration files and instructions for reproducing all experiments reported in the paper. 

We used large language models (LLMs) in a limited capacity to assist with the writing of this paper and the design of parts of the code. Specifically, LLMs were employed only to (i) improve sentence clarity and conciseness, (ii) condense overly lengthy paragraphs, and (iii) provide coding assistance for implementation design. All technical contributions — including the method design, experimental setup, results, analyses, and final implementation decisions — are entirely our own work.

\end{document}